# SAPPO
# Sistema de Alta Precisión de POsicionamiento

Sistema de posicionamiento en interiores

# SAPPO
# High Precision Positioning System

Indoor positioning system

Dr. Antonio Losada González
anlosada@uvigo.es
Universidad de Vigo

## 1. Abstract

SAPPO is a high-precision, low-cost and highly scalable indoor localization system. The system is designed using modified HC-SR04 ultrasound transducers as a base to be used as distance meters between beacons and mobile robots. Additionally, it has a very unusual arrangement of its elements, such that the beacons and the array of transmitters of the mobile robot are located in very close planes, in a horizontal emission arrangement, parallel to the ground, achieving a range per transducer of almost 12 meters. SAPPO represents a significant leap forward in ultrasound localization systems, in terms of reducing the density of beacons while maintaining average precision in the millimeter range.

## 2. Resumen

SAPPO es un sistema de localización en interiores de alta precisión, bajo coste y alta escalabilidad. El sistema está diseñado empleando como base transductores de ultrasonido HC-SR04 modificados para ser empleados como medidores de distancia entre balizas y robot móvil. Adicionalmente presenta una disposición muy poco común de sus elementos, de forma que las balizas y el array de emisor es del robot móvil se encuentran ubicados en planos muy cercanos consiguiendo un alcance por transductor de casi 12 metros. SAPPO presenta un salto importante dentro de los sistemas de localización por ultrasonidos, en cuanto a la reducción de la densidad de balizas manteniendo una precisión media en el rango milimétrico.

## 3. Análisis actual de los sistemas de localización en interiores por ultrasonidos

Entre los sistemas de localización por ultrasonidos, tenemos gran parte de ellos que emplean sus mediciones para realizar medidas relativas a los objetos más próximos, trabajos de estos tenemos en [8] [9] [10] [11] [12] [13] [14] [15] [16] [17] [18] [19] [20] [21] en los que las señales de ultrasonidos se emplean como sonar para mapear su entorno. Este tipo de sistemas solo comparten la necesidad de medir el tiempo de vuelo de la señal de ultrasonidos con los sistemas de localización global, por lo que no serán analizados más trabajos de este tipo en los siguientes apartados.

La inmensa mayoría de los trabajos revisados están diseñados para poder ser empleados de modo universal en cualquier instalación obteniendo una ubicación absoluta, por lo que sitúan las balizas en el techo de los habitáculos como en [22] [3] [4] [1] [23] [24], calculando la distancia entre las balizas y la estación móvil que se encuentra en el suelo, necesitando un mínimo de tres medidas para obtener una localización. Esta disposición aplicable a cualquier instalación, presenta el inconveniente de tener una muy baja cobertura y no aprovecha la capacidad máxima de emisión lineal de los transductores de ultrasonidos. Cada transductor, incluso en los tipos más económicos, es capaz de cubrir una distancia lineal de 8 a 12 metros y su cono de radiación alcanza los 30 grados. Empleando la disposición más habitual con la colocación en el techo de las balizas (Como en el caso del sistema Constellation que puede verse en la Figura 1), en una casa estándar cuya altura es de 2,50 m, dado que es el mínimo legal en la actualidad, la cobertura no supera los 1,5 m2 en el suelo, aunque presenta como ventaja, el tener la certeza de que no encontraremos obstáculos entre las balizas y los dispositivos a localizar.

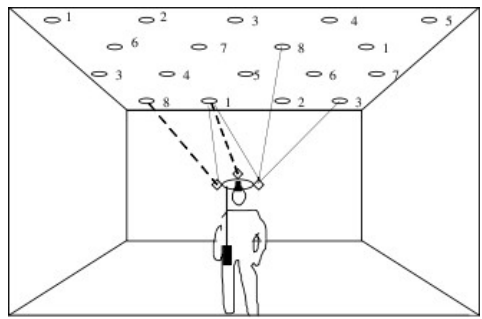

Figura 1. Distribución de balizas en el ILS Constellation.

En otro tipo de disposición muy poco explorada y sobre la que no hemos encontrado ningún trabajo con implantaciones en localizaciones reales, se sitúan los emisores a cierta altura, con su cono de emisión paralelo, o casi paralelo al suelo como en [25] [26] [27] [28]. En la Figura 2 puede verse esta disposición, con una baliza emitiendo su señal con el centro del cono de emisión formando un ligero ángulo con el suelo.

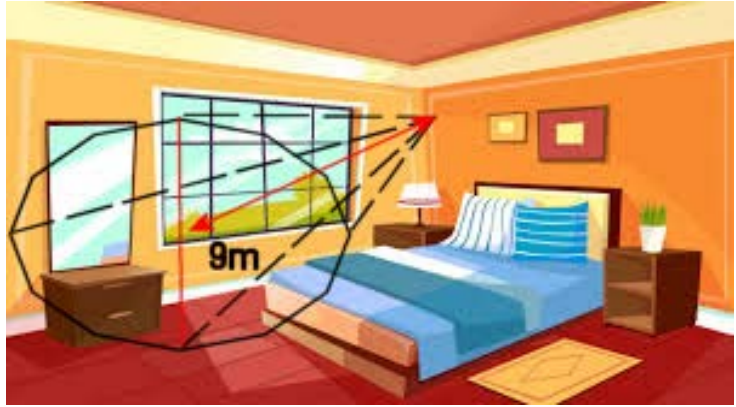

Figura 2. Cobertura máxima de una baliza en SAPPO.

La mayor parte de los trabajos analizados se centran en describir sistemas de localización muy precisos empleando distintas tecnologías dentro del ámbito de los ultrasonidos, pero en casi ninguno de ellos se realizan estudios de complejidad de despliegue en entorno real ni estudios precisos del área de cobertura que posibilita la localización en despliegues en entornos reales, ni de ubicaciones, ángulos de inclinación y colocación de las balizas.

Gran parte de los trabajos se realizan en entornos de laboratorio, donde se colocan las balizas en el techo orientadas con cierto ángulo hacia el suelo donde se sitúa la estación móvil a localizar como en [7] [6]. En estos casos, el área de cobertura real sería la resultante del solapamiento de los conos de radiación ultrasónica de todas las balizas, que normalmente es bastante pequeña y está centrada en la habitación como puede apreciarse en la Figura 2.

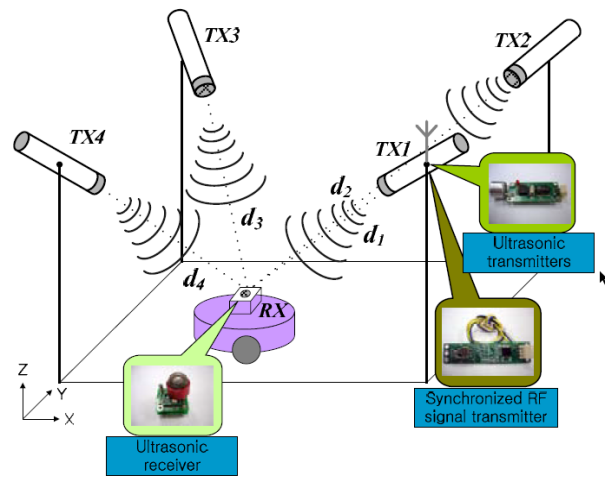

*Figura 3. Cobertura de balizas situadas en el techo..*

La función desempeñada por las balizas dentro de los sistemas de localización por ultrasonidos condiciona el comportamiento y las características del sistema, éstas pueden encargarse de emitir el pulso de ultrasonidos, recibirlo, o incluso poseer capacidad para ambas funciones.

Normalmente las balizas son siempre emisoras y los robots móviles son los que reciben el pulso de ultrasonidos, de esta forma es el robot el que tiene la información para realizar el cálculo de su posición.

En un sistema de localización con balizas es posible asignar la función de emisión y recepción tanto al robot móvil como a las balizas. En los trabajos en los que se realiza una localización distribuida como en [30] [28] [31], todos los sensores tienen la capacidad de emisión y recepción, dado que tanto las balizas como los robots son puntos de referencia. En caso de que las balizas sean las que emitan el pulso de ultrasonidos y el robot sea el receptor, normalmente tenemos un sistema de alto consumo debido a su emisión constante, pero muy escalable en lo relativo al número de robots a localizar dado que los robots son receptores pasivos, como en el sistema GPS global.

Dentro de este comportamiento se engloban todos los sistemas de emisión de banda ancha, dado que aprovechan la capacidad de emisión simultánea de todas sus balizas, como los mostrados en [32] [33] [34] [35].

En el caso de emplear codificación de señal en banda estrecha, para el acceso al medio por las balizas para la emisión del pulso de ultrasonidos, puede emplearse TDMA como en [36] [37] [38], en el que cada baliza emplea su slot de tiempo para transmitir, con lo que tendremos un ligero incremento de la imprecisión originada por el movimiento de robot, dado que la velocidad de los ultrasonidos es relativamente lenta, la medida de cada baliza se realizará en un instante de tiempo diferente, lo que incorporará un error al combinarlas para calcular la posición del robot debido a su desplazamiento, aunque este error puede ser reducido estimando la posición en cada una de las lecturas.

En otros trabajos como [23] el robot envía un pulso de radio activando las balizas de modo secuencial. El problema con este sistema es su lentitud en despliegues grandes, debido al alto número de balizas.

En el caso contrario, en que sea el robot el que transmita el tren de pulsos de ultrasonidos, las balizas no consumirán casi energía mientras no les llegue el paquete de radio de sincronización solicitando una medición, con lo que mantendrán un consumo de energía reducido. Tampoco tendremos el problema de latencia o el error producido por el movimiento de los robots ni siquiera en sistemas de banda estrecha, dado que el robot móvil transmitirá al mismo tiempo el tren de pulsos a todas las balizas a su alcance. Pero aparecen otros problemas, dado que las balizas son las que reciben el pulso de ultrasonidos, son éstas las que miden el tiempo de vuelo que necesita el robot para calcular su posición, con lo que después de la medida hay que diseñar un procedimiento para que transmitan esta información al robot o a un nodo central para realizar los cálculos de posición.

Ejemplos de este modo de funcionamiento pueden verse en [26] [4] [5]. Este modo no se aconseja para entornos cooperativos de robots, dado que todos ellos estarían solicitando sus medidas a la vez, lo que complicaría mucho el sistema, dado que tendría que diseñarse un sistema que evite las posibles colisiones.

Dentro de los trabajos de investigación que encontramos sobre sistemas de localización empleando ultrasonidos tenemos varios tipos de codificación de la señal. Como clasificación inicial, la información enviada por los transductores de ultrasonidos puede estar codificada en banda estrecha o en banda ancha.

Se codifica en banda estrecha cuando los transceptores emiten en una única frecuencia. Una de las limitaciones más importantes es la imposibilidad de transmisión simultánea de varias balizas. Para poder emplear el mismo medio, las balizas deben sincronizarse entre sí repartiendo el canal en slots de tiempo mediante la técnica TDMA (Time Division Multiple Access).

Mediante este tipo de acceso al medio, los transductores de las balizas envían un tren de pulsos de ultrasonidos solo en el slot de tiempo que le corresponde, que son recibidos por los transceptores destino como en [30] [39] [31] [40]. En casi todos los trabajos de este tipo, los transductores emiten una portadora de 40 KHz que se considera un estándar en emisión de ultrasonidos. Un caso particular es el mostrado en [41] [2] donde se emplean ultrasonidos de banda estrecha, pero en este caso se emplean distintos tipos de transductores, cada uno diseñado para una frecuencia distinta. Esta aproximación permite que todas las balizas transmitan su pulso en banda estrecha a la vez sin interferencias, pero complica el hardware, dado que la estación móvil debe tener uno o varios receptores para cada tipo de frecuencia.

Otra parte de los trabajos emplean codificación de la señal empleando espectro ensanchado mediante CDMA (Code Division Multiple Access) como [42] [3] [43], lo que permite que múltiples balizas actúen como emisoras de modo simultáneo. Dentro de esta tecnología los sistemas de ultrasonidos emplean FHSS o DSSS y en algún trabajo han empleado OFDM. Ejemplos de sistemas de este tipo se muestran en [44] [3] [45] [46] [27] [47] [48].

En cuanto al sistema de cálculo de las medidas a emplear para obtener la localización hay dos opciones, puede medirse el tiempo de vuelo de la señal de ultrasonidos entre la baliza y el robot o puede medirse el ángulo de incidencia, incluso en trabajos como en [23] se mide el tiempo de vuelo a un sensor con 3 receptores equidistantes en un plano y con esta información se calcula el ángulo de incidencia.

Entre los sistemas que emplean el tiempo de vuelo de la señal, hay dos tipos, los que emplean el tiempo de vuelo de cada una de las balizas independiente y los que calculan la diferencia de tiempo de vuelo entre las señales de todas las balizas.

En los sistemas que emplean el tiempo de vuelo de cada baliza de forma independiente, conociendo el tiempo de vuelo y la velocidad de propagación del sonido en el aire, se calcula de forma muy precisa la distancia a cada una de las balizas. Mediante trilateración, obteniendo tres medidas de distancia a tres puntos conocidos se puede calcular la intersección de las esferas generadas por cada una de las medidas de distancia permitiendo calcular la posición del robot. En [52] [51] [49] [50] pueden encontrarse trabajos que emplean esta técnica. En estos sistemas las balizas deben estar perfectamente sincronizadas entre sí y con el robot.

En otro tipo de sistemas como [53] [54] [6] [55] [56] [57] [58] [59] se emplea la diferencia del tiempo de vuelo. En este caso no se tiene el tiempo de vuelo de cada baliza de forma independiente, sino que se obtiene el tiempo transcurrido entre la recepción de los ultrasonidos de cada baliza. En estos sistemas, las balizas tienen que estar sincronizadas entre sí, pero no con el robot. En este caso deben emplearse cálculos hiperbólicos para conocer la ubicación del robot.

Para localizar el robot también pueden emplearse algoritmos de triangulación. En este caso es necesario conocer el ángulo de incidencia de las emisiones de ultrasonidos de tres balizas sobre el robot. Ejemplos de esta técnica se encuentra en los trabajos [60] [25] [61] [23] [46]. Adicionalmente, en algún trabajo se emplea la triangulación mediante la obtención de los ángulos de incidencia de las ondas de ultrasonidos mediante el cálculo de la diferencia de fase con la que inciden las ondas en dos sensores próximos, pero estos trabajos ofrecen precisiones muy bajas por debajo del rango exigible a este proyecto.

## 4. Introducción

SAPPO, cuyo logo se muestra en la Figura 4, es un sistema de posicionamiento en interiores, que permitirá ubicar robots móviles en cualquier lugar de una vivienda, empleando señales de ultrasonido en banda base a 40 KHz.

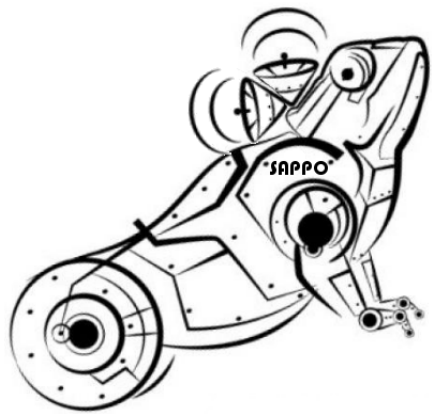

*Figura 4 Logo del sistema de posicionamiento SAPPO.[1]*

Para que esto sea posible, estará compuesto por un conjunto de balizas y un dispositivo situado en el robot móvil que se quiera localizar. Las balizas estarán ubicadas en puntos fijos conocidos y estarán repartidas por toda la casa. Para su funcionamiento sólo será necesario la instalación del cable de alimentación, ya que las comunicaciones se realizarán de modo inalámbrico. Uno de los objetivos principales de todo el sistema es limitar al mínimo el número de balizas necesarias para conseguir la localización en toda la casa, o al menos en todas las áreas en las que se encuentren caminos de circulación que pueda seguir el robot para desplazarse de una habitación a otra.

Cada uno de los robots móviles a localizar, deberán disponer de un dispositivo electrónico que le permita comunicarse con las balizas y recuperar la información de distancia a cada una de las balizas alcanzables desde el dispositivo en cada una de las habitaciones. Este dispositivo será el encargado de realizar los cálculos necesarios para determinar la posición del robot.

## 5. Función de cada elemento

El sistema de localización está compuesto por balizas y por robots móviles. Inicialmente podría seleccionarse como emisor o receptor a cualquiera de los elementos, pero la elección asignará conjuntos de características distintas al sistema.

El asignar el rol de recepción a los robots móviles y el emisor a las balizas nos permite generar un sistema escalable que permite la coexistencia de varios robots en la misma habitación, pero el

---
[1] Fuente: https://dibujosa.com/dibujosgratisapp.php?codigo=18806 [Acceso: 4-abr-2022]

inconveniente, al haber seleccionado una portadora de banda base a 40 KHz, es que no pueden transmitir todas las balizas a la vez, de modo que siempre habrá una pequeña diferencia de tiempo entre la transmisión de cada baliza, que generará imprecisiones cuando el robot se encuentre en movimiento. Adicionalmente, el retraso en las medidas ocasionada por este retardo generará un menor número de medidas por unidad de tiempo. En este caso, debemos tener en cuenta que este retardo entre transmisiones no será despreciable, dado que entre cada medida tiene que transcurrir un tiempo mínimo para que la energía del tren de pulsos de ultrasonidos se disipe. Teniendo en cuenta el alcance de los transductores empleados y calculando la distancia máxima que puede llegar a recorrer el ultrasonido, dado que no queremos que genere falsas detecciones positivas en la siguiente medida, debemos esperar un mínimo de 3 centésimas de segundo.

Por los anteriores motivos, se ha seleccionado como emisor al robot móvil y como receptores a las balizas, de modo que el emisor enviará un paquete de radio de sincronización global para todas las balizas y después enviará un pulso ultrasónico omnidireccional que detectarán todas las balizas a la vez, con lo que no se generarán retardos en movimiento. El paquete de radio que envía el robot móvil identifica el robot que solicita la medida, de modo que pueden coexistir varios robots a la vez en el mismo espacio, siempre que no soliciten medidas de posicionamiento de modo simultáneo.

El sistema elegido tiene inicialmente dos problemas, por un lado, es poco escalable, dado que una misma emisión de las balizas no puede ser empleada para localizar a todos los robots a la vez, pero por otra parte debemos tener en cuenta que este sistema está diseñado para ser empleado en el interior de casas, por lo que será muy poco habitual que dentro de una vivienda particular existan varios robots que necesiten emplear el sistema de localización, por lo que no es necesario crear un sistema escalable que permite localizar varios dispositivos de modo simultáneo.

Adicionalmente, este sistema genera un segundo problema, dado que el robot móvil envía el paquete de sincronización de radio con información del robot que solicita la medida, y son cada una de las balizas las que miden el tiempo de vuelo de las ondas de ultrasonido, la información de los tiempos de vuelo la tiene cada una de las balizas de modo independiente, por lo que será necesario transmitir toda esta información al robot móvil para que este pueda calcular su posición. Dada la emisión múltiple de información simultánea, en este punto deben controlarse las posibles colisiones.

Se ha escogido esta configuración dado que tiene como ventajas, la precisión de la localización, e incrementa notablemente el número de medidas que pueden efectuarse por segundo, en detrimento del número máximo de robots que puede gestionar en una misma habitación.

# 6. Descripción de las balizas

## 6.1. Ubicación de las balizas para mejorar su cobertura

Este apartado está íntimamente relacionado con uno de los requisitos principales del sistema. Se debe tener en cuenta que el objetivo de la tesis es desarrollar un sistema real que pueda ser implantado en una vivienda estándar, y para que la implantación no sea muy compleja, debe requerir la instalación del menor número de balizas posible, dado que cada baliza debe ser cableada al menos con la alimentación. Para los lugares donde resulte especialmente complicado cablear una baliza puede emplearse un hardware específico de muy bajo consumo que permite periodos largos de recambio de baterías, pero este sistema debe ser empleado como medida excepcional, dado que será necesario revisar y cambiar sus baterías cuando se agoten, lo que complica el mantenimiento del sistema.

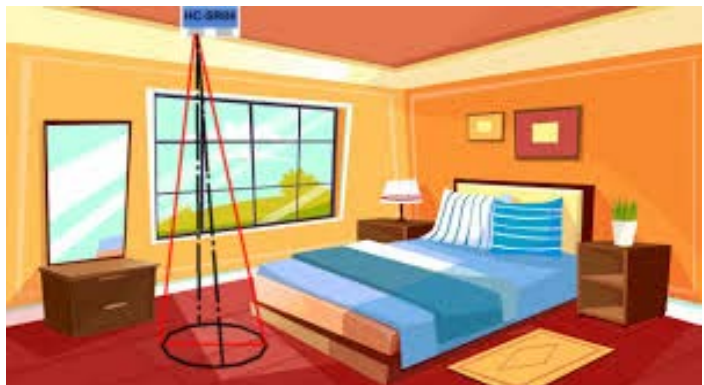

*Figura 5 Cobertura de baliza instalada en el techo.*

En la mayor parte de los sistemas de posicionamiento por ultrasonidos, las balizas se sitúan en el techo, como se puede apreciar en la Figura 5, en donde se ha dibujado solo una baliza en el techo orientada hacia el suelo. El área cubierta por cada baliza aumenta notablemente con la distancia, pero con esta disposición estamos limitados por la altura de la planta. Resulta evidente que cuanto más lejos se encuentren las balizas del dispositivo a posicionar se tendrá un mayor rango de cobertura por baliza, siempre que nos mantengamos dentro del alcance físico. En la Figura 5 podemos ver el área de cobertura de un solo sensor sobre el suelo, que normalmente se ve todavía más reducida dado que el receptor del robot se encuentra a cierta altura del suelo.

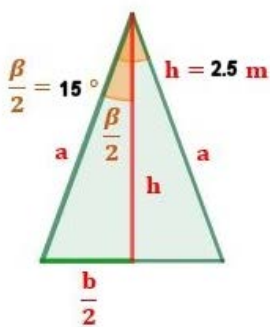

*Figura 6.Cono de emisión ultrasónico.*

Para calcular el área cubierta por un transceptor podemos simplificar el cono de radiación pasándolo a 2D como podemos ver señalado en la Figura 6 en líneas rojas y convertirlo en un triángulo isósceles como sigue: la mayoría de los transceptores de ultrasonidos tienen un ángulo de emisión de 30 grados y teniendo en cuenta que la suma total de los ángulos del triángulo que forma la radiación es de 180 grados, cada uno de los ángulos inferiores es de 75 grados.

$$\tan\frac{\beta}{2} = \frac{\frac{b}{2}}{h} => b = \tan\frac{\beta}{2} \cdot 2h \tag{1}$$

Siguiendo la Ecuación (1) y teniendo en cuenta una altura del techo de 2,5 m, obtenemos un círculo de 1,34 metros de diámetro de cobertura de la radiación en el suelo, lo que representa un área de

$$\pi \cdot r^2 = \pi \cdot \left(\frac{1,34}{2}\right)^2 = 1,41\ m^2 \tag{2}$$

Como puede apreciarse, el área cubierta por cada baliza con un único transceptor de ultrasonidos es extremadamente pequeña, siendo necesario desplegar un número muy alto de balizas para cubrir una única habitación.

Teniendo esto en cuenta y sabiendo que el sistema debe ser empleado en una casa estándar, debe tenerse en cuenta las características de una casa a sus distintas alturas. Si queremos alcanzar el mayor grado de cobertura debe cambiarse la ubicación de los sensores, en vez de estar ubicados en el techo y orientados hacia el suelo, estarán ubicados a una altura de 1,70 m y orientados paralelamente al suelo. Esta orientación no podría ser aplicada de modo universal, sin embargo, es una configuración perfectamente adaptada a las disposiciones de los objetos en la mayoría de las viviendas. Para que esta disposición sea válida, el cono de radiación de las balizas no debe encontrarse con ningún objeto en su trayecto. Reflexionando sobre la ubicación de los objetos con altura superior a 1,70 m en una vivienda, encontramos que todos los objetos altos se ubican en las paredes, estando el centro de las habitaciones libre a esa altura, por lo que no encontramos obstáculos a la altura indicada. Con esta disposición podremos cubrir las habitaciones de modo longitudinal aprovechando la capacidad de emisión lineal de cada uno de los transceptores, dado que normalmente las habitaciones son más largas y anchas que altas.

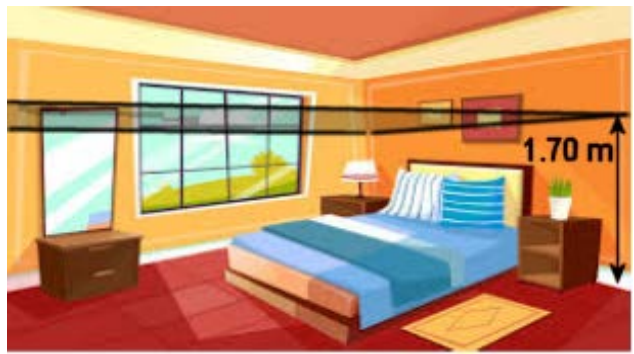

*Figura 7.Área de colisión a 1,70 m de altura.*

Como podemos ver en la Figura 7, a 1,70 m no encontramos obstáculos en el centro de las habitaciones, con lo que las ondas de ultrasonidos no colisionarían con ningún objeto, por lo es posible obtener medidas directas.

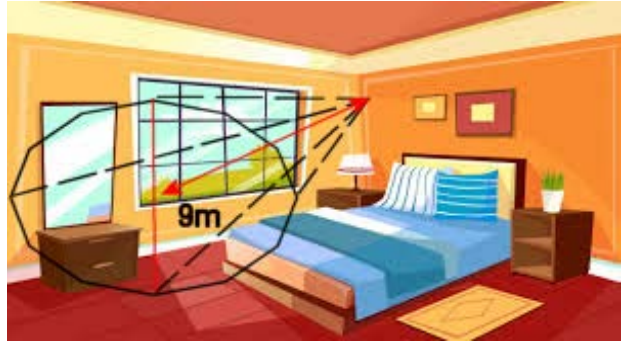

*Figura 8. Volumen cubierto por el cono de emisión de una baliza en SAPPO.*

En esta Figura 8 puede apreciarse una simulación del área de radiación de un solo transceptor de ultrasonidos. en la que puede apreciarse el incremento del área de cobertura mediante el cambio de orientación.

Con el cambio de orientación y calculando de nuevo el triángulo aproximado resultante de la zona de cobertura, pero ahora con una altura de 9 m, que es la distancia media máxima que alcanzan la mayoría de los sensores de ultrasonidos, tenemos una superficie de cobertura de 18 m² que es un 1300% de la superficie cubierta con las balizas en el techo.

Debemos tener en cuenta que el área de proyección ya no es el área del círculo de proyección en el suelo, sino el área de todo el triángulo del cono de radiación proyectado a 9 metros de distancia de la baliza.

En la Figura 8 podemos ver la primera planta de la casa que se ha empleado para probar el prototipo del sistema. En la misma se puede apreciar la proyección sobre el suelo del área de cobertura de un solo transceptor de ultrasonidos sobre el salón, que tiene una longitud de 10 metros.

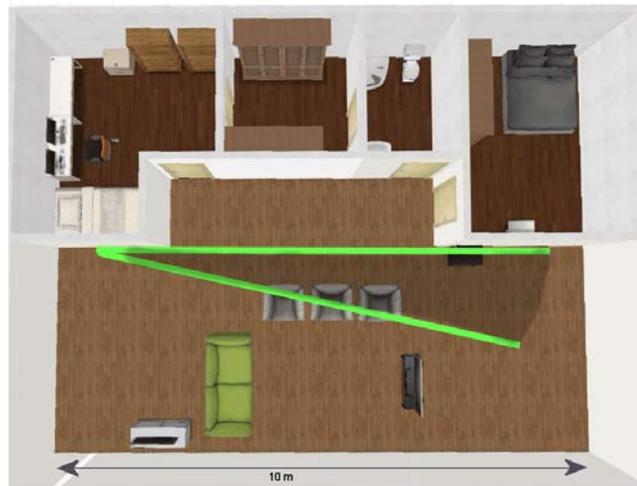

Figura 9. *Proyección en suelo de cobertura de un transceptor de ultrasonidos.*

Aun así, podemos ver que el espacio sin cubrir sigue siendo demasiado grande. Debemos tener en cuenta que cada transceptor contiene solo la electrónica de generación de portadora de ultrasonidos más los módulos amplificadores de señal y los de decodificación de lectura para leer los pulsos de ultrasonidos. Estos transceptores deben estar controlados por un microcontrolador para recuperar la información de las transmisiones. El conjunto de los transceptores de ultrasonido, la electrónica de potencia y el microcontrolador se denomina baliza y deben tener conexión con la red eléctrica.

Pero en ningún caso estamos limitados a que una baliza contenga solamente un transceptor de ultrasonidos. Dado que cada baliza debe estar cableada a la red eléctrica, debemos limitar el número de balizas a desplegar y para ello se ha incorporado a cada una de ellas un array de 4 transceptores de ultrasonidos para cubrir una zona de 90 grados. Con esta nueva configuración, cada una de las balizas cubre un área circular de 90 grados de arco y 9 metros de distancia, cuyo centro es la propia baliza, no habiendo ninguna limitación que impida unir dos arrays de 4 transceptores y construir una única baliza con 8 transceptores para cubrir un arco de 180 grados y 9 metros de radio o incluso colocar 16 transceptores para cubrir los 360 grados.

En la Figura 9 se puede ver en amarillo el área de cobertura que se obtiene con una sola baliza con 4 transductores.

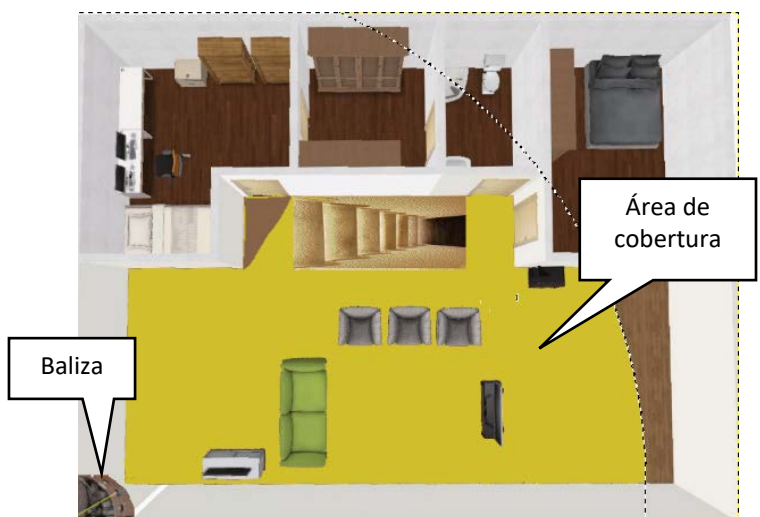

*Figura 10.Área cubierta por una baliza con 4 transductores.*

Con esta nueva disposición, una sola baliza con un arco de cobertura de 90 grados cubriría 63.59 m² llegando a cubrir hasta 254 m² teóricos en espacio libre, ampliando el número de transductores de la baliza hasta cubrir los 360 grados.

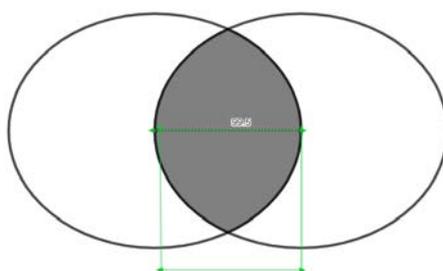

*Figura 11.Área de localización con dos balizas instaladas en el techo.*

Como muestra la Figura 10, para conocer la cobertura real del sistema debemos calcular la intersección de las áreas de cobertura de dos balizas, dado que es el número mínimo de medidas que necesitamos tomar para triangular una posición en el plano.

$$A = \left(\frac{2}{3}\pi - \frac{\sqrt{3}}{2}\right)r^2 \qquad (3)$$

La Ecuación (3) representa el área de intersección entre las áreas de cobertura de las dos balizas. En el caso propuesto en el que estamos empleando balizas con cobertura de 90 grados de arco, el área de cobertura por las dos balizas es la mitad que el área de cobertura total y en este ejemplo asumimos que la distancia entre balizas es igual a la distancia máxima de cobertura que son 9 m. Con estos datos obtenemos que el área máxima que pueden cubrir dos balizas distanciadas 9 m con 4 transceptores cubriendo un arco de 90 grados es de 49,74 m2 que corresponde a la mitad del área de cobertura total. En la Figura 11 podemos ver la intersección de dos círculos y podemos apreciar encuadrada en color verde la zona que correspondería a la habitación y en color gris la zona de la intersección del área de emisión de estas dos balizas con arco de cobertura de 90 grados. En el ejemplo, la habitación tendría 81 m² siendo un tamaño excesivo. Cuanto menor sea la distancia entre las balizas y más corta la habitación, el área de intersección cubrirá un mayor porcentaje de la habitación. Con esta nueva disposición las limitaciones normalmente vienen marcadas por las posiciones de las paredes, dado que los sensores que podemos emplear oscilan entre 9 y 14 metros de cobertura lineal y difícilmente podremos encontrar en una casa estándar una habitación de más de 14 m de largo.

En nuestro sistema es posible realizar el cálculo de la posición con solamente dos balizas, dado que conocemos la altura a la que están situadas las balizas y conocemos la altura del receptor.

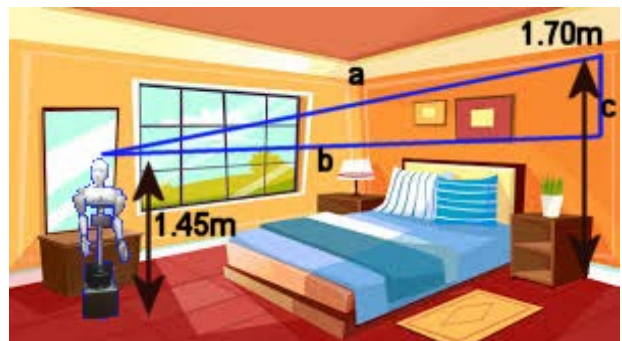

*Figura 12.Triángulo de ajuste de posición con balizas instaladas a 1,70 m de altura.*

Como podemos ver en la Figura 12 el emisor del robot se encuentra a 1,45 m y la baliza a 1,70 m. Aunque el emisor del robot no esté situado en el plano de las balizas, conociendo la altura del sensor del robot y la altura de las balizas podemos obtener **C**, y podemos obtener **A** calculando la distancia entre la baliza y el sensor del robot mediante el sistema de posicionamiento, con estas dos medias podemos calcular fácilmente **B**, ya que todas las medidas forman un triángulo rectángulo. Resolviendo **B** tenemos una de las medidas necesarias para triangular la posición en el suelo. Con el mismo procedimiento debemos obtener otra distancia desde otra baliza posicionada en un lugar distinto para obtener una ubicación precisa.

Para obtener una localización en dos dimensiones necesitamos un mínimo de 2 medidas de distancia a dos balizas siempre que, por disposición de las balizas, uno de los dos puntos de intersección de las circunferencias de cobertura de cada una de ellas corresponda con valores imposibles por no encontrarse en el habitáculo o podamos descartarlo mediante medidas de posiciones cercanas o por el cálculo de la trayectoria. Al necesitar dos medidas, la superficie total de cobertura estará compuesta por el área de intersección de ambas balizas. Adicionalmente hay que tener en cuenta que las balizas deberán recibir una señal de ultrasonidos, pero esta señal debe ser emitida por el robot móvil, por lo que es necesario tener en cuenta la cobertura del sistema instalado en el robot. Finalmente, el área final total de cobertura será la intersección de un mínimo de dos balizas junto con el robot, que puede quedar reducido a la intersección de las dos balizas, siempre que el robot comparta el mismo rango de cobertura.

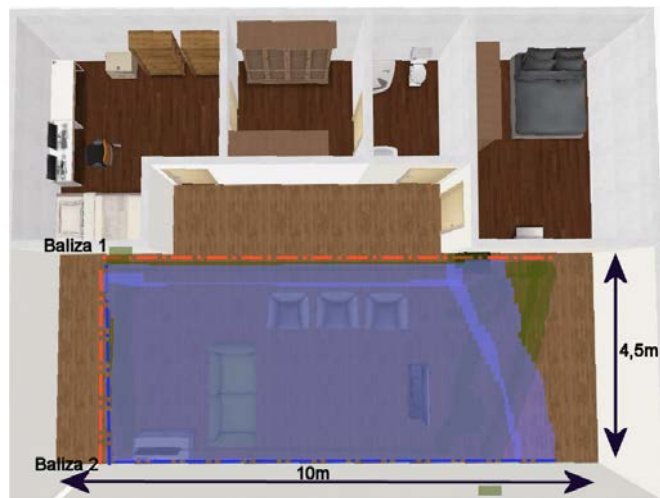

*Figura 13.Área de localización con balizas con arrays de transductores instaladas paralelas al suelo.*

En la Figura 13 puede verse el gráfico de intersección de las áreas de cobertura de las balizas en una habitación real de grandes dimensiones. En la Figura 14 podemos ver una representación gráfica del cálculo de cobertura en una habitación de una planta de una vivienda, en donde entran en juego las paredes (en verde) para calcular la cobertura total. En esta representación, para que sea más gráfica, se ha asumido que la anchura de la habitación es igual a la distancia máxima de cobertura de la baliza (9 metros en los transceptores más económicos) y que la longitud de la habitación supera por muy poco esta distancia máxima de cobertura. En esta situación ideal de máxima cobertura con dos balizas es de 38 m².

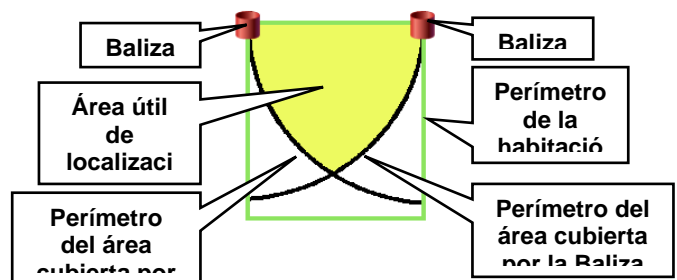

*Figura 14.Área de intersección de las balizas.*

En el Anexo II puede verse de modo detallado el algoritmo de cálculo empleado para calcular la cobertura real de este sistema en una habitación de 4,5 m de anchura y 11,5 m de longitud.

Dado que los transceptores de ultrasonidos son directivos y el robot debe ser capaz de emitir ultrasonidos que alcancen a cualquiera de las balizas, independientemente a su orientación, el robot debe montar un array de transceptores de ultrasonidos. Teniendo en cuenta su rango de cobertura de 30 grados por transceptor para todo el alcance, debe poseer un anillo circular de 12 transceptores para cubrir los 360 grados, de este modo, independientemente a su orientación, será capaz de tener a su alcance las balizas.

Mediante esta disposición, siempre que respetemos las condiciones, podremos emplear solo dos balizas en cualquier habitación rectangular. Siempre que no supere los 9x9 metros podremos emplear transceptores muy económicos y en caso de que los supere podemos alcanzar los 14x14 metros con transceptores de mayor coste. Esta disposición nos permite reducir al mínimo el número de balizas a desplegar con lo que reducimos los tiempos de despliegue, la necesidad de instalación de cableado y el coste del despliegue total.

## 6.2. Estimación de densidad de balizas

Dada la altísima cobertura con solo dos balizas, para calcular el área de cobertura en una habitación real será necesario calcular el área teniendo en cuenta las limitaciones impuestas por las paredes. En el caso de la planta de la casa en donde se realizaron las pruebas de desarrollo del sistema, empleando dos balizas de las más económicas con rango de cobertura de 9 metros, tenemos una cobertura real de 37,4 $m^2$ en una habitación muy grande con una superficie de 45 $m^2$. En el Anexo II se muestra el procedimiento completo para realizar el cálculo de cobertura de la planta.

## 7. Codificación de la señal

De entre todas las posibilidades se ha seleccionado la codificación de la señal de ultrasonidos empleando una portadora en banda base de 40 KHz, dado que existe multitud de hardware muy económico disponible, empleando este sistema de codificación. Los dispositivos transductores con portadoras de 40 KHz están muy probados y dada la sencillez de la portadora tienen una muy alta velocidad de detección de la señal, lo que incidirá en la posibilidad de aumentar la frecuencia de muestreo, y mejorar la precisión de las medidas, que será necesaria para mantener la precisión al localizar robots en movimiento.

En el Anexo III enumeramos los transceptores disponibles comercialmente que emplean portadoras simples de 40 KHz. Disponemos de varios modelos con características diferentes en cuanto a coste, cobertura lineal, ángulo de apertura de la señal y posibilidades de configuración y programación. Físicamente los modelos también son diferentes poseyendo dos configuraciones, los que emplean un mismo transductor para emitir y para recibir la señal y los que emplean uno para cada función.

## 8. Sistema de sincronización

Al emplear transductores con portadora simple de 40 KHz no es posible enviar información en la señal, ni emplear la señal como mecanismo de sincronización, ni tampoco es posible transmitir con dos transceptores de modo simultáneo desde dos ubicaciones diferentes en el mismo espacio de cobertura, ya que sus ondas de sonido provocarían interferencias entre sí, no siendo posible detectar ninguna de ellas correctamente, y en caso de ser detectadas no sería posible identificar su origen.

Por este motivo la sincronización es necesaria para el funcionamiento del sistema de medición y debe ser realizada de modo independiente a la propia señal.

De entre las distintas alternativas, se ha empleado un mecanismo de sincronización por transmisión de ondas de radiofrecuencia. Antes del inicio de un ciclo de localización, el robot móvil enviará simultáneamente el tren de pulsos de ultrasonidos y un paquete de sincronización de radio que será recuperado por todas las balizas, momento en el que todas iniciarán el contador de tiempo para calcular el tiempo de vuelo de la señal de ultrasonidos. Cuando detecten el tren de pulsos de la portadora de ultrasonidos, pararán el contador de tiempo y almacenarán el tiempo medido para enviarlo posteriormente al robot. Una vez que el robot recupere las medidas de las balizas suficientes podrá triangular su posición. El tiempo de vuelo del paquete de radio se considera nulo dada la altísima velocidad relativa de las ondas de radio con respecto a las ondas de sonido, con lo que se puede despreciar. Uno de los puntos a tener en cuenta en este sistema de sincronización es el sistema de detección de los paquetes de radio, dado que, aunque el tiempo de vuelo de las ondas electromagnéticas es despreciable en distancias muy cortas, la codificación, decodificación y detección del paquete de radio, si es un factor a tener en cuenta en caso de estar trabajando con precisiones de localización muy altas.

## 9. Descripción del hardware

Tanto en las balizas como en el robot es necesario disponer de cuatro sistemas, el sistema de sincronización, el sistema de comunicación, el sistema de medida y el sistema de cómputo.

### 9.1. Transductores de ultrasonidos SRF04. Modificación de sensores para el empleo en sistemas de posicionamiento

El hardware de ultrasonidos está compuesto por los transductores de ultrasonidos. Tanto el robot como las balizas contienen un array de transductores de ultrasonidos, por lo que será necesario escoger un dispositivo de bajo coste, ya que este coste se multiplicará por el número de componentes del array.

En el anexo III se enumeran los distintos transceptores comerciales actuales. Entre los disponibles al inicio del desarrollo de este proyecto estaba el SRF02 y el SRF04. El SRF02 tenía un coste muy alto de 30€ por unidad lo que incrementaría el coste del hardware en exceso. Aunque estos dispositivos tienen una cobertura muy alta de 12 metros y pueden operar tanto como emisores como receptores o como detectores de obstáculos, tienen un coste del orden de 30 veces superior a los transductores seleccionados.

La alternativa son transceptores SRF04. Estos módulos son muy económicos, con un coste aproximado de 1 € por unidad. Tienen una cobertura de 4,5 m para detección de obstáculos, en donde la onda ultrasónica realiza el camino de ida y vuelta, que se transforma en una cobertura lineal de 9 metros con una precisión media inferior a 2 cm. El inconveniente principal es que solo está pensado para funcionar como detector de obstáculos. En este modo de operación el sensor envía un pulso de ultrasonidos y espera hasta que recibe de nuevo la señal de retorno. El tiempo transcurrido es el tiempo de vuelo de la señal de ultrasonidos. Al ser un sensor muy económico, el tiempo debe contabilizarse con un sistema externo.

En el SRF04, la operación de medida de tiempo siempre comienza con el envío del pulso de ultrasonidos. Como no es posible evitar el envío del pulso ultrasónico, en su configuración original no es posible emplear un SRF04 como emisor y otro SRF04 como receptor, dado que el receptor también enviará un pulso de ultrasonidos.

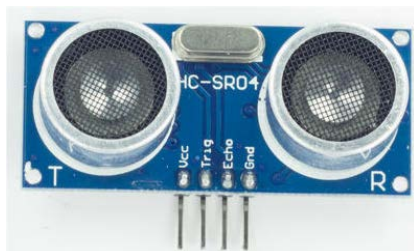

*Figura 15.Imagen de un sensor de distancia SRF-H04.*

En la Figura 15 se ve una imagen del módulo de bajo coste HC-SR04 que es una de las copias compatibles con el SRF04. Este módulo solo tiene 4 pines, los dos de alimentación positivo y negativo, echo y trigger. El pin trigger es el disparador de los pulsos de ultrasonidos y el pin de echo se pone a nivel alto en cuanto se envía el tren de pulsos de ultrasonidos y se pone a nivel bajo cuando detecta el tren de pulsos de vuelta, una vez que rebota en algún objeto. Estos sensores de bajo coste no están diseñados para poder ser empleados dentro de un sistema de posicionamiento, dado que el mismo sensor realiza la emisión y la recepción del pulso ultrasónico. Para posibilitar el empleo de este transceptor únicamente como un receptor hay que eliminar o deshabilitar la electrónica que realiza la función de envío del tren de pulsos. Con ese objetivo, simplemente ha sido necesario localizar y desoldar el emisor de pulsos. De esta forma, la electrónica del transductor realizará todas las etapas para enviar el tren de pulsos de ultrasonidos, pero dado que se le ha desconectado el emisor de ultrasonidos estos no podrán generarse y enviarse al aire, y pasará a modo recepción. Con esta pequeña modificación reconvertimos los transceptores para operar en modo lectura. Teniendo en cuenta que entre las balizas y el robot suman 20 transceptores, empleando estos sensores mantenemos el coste de los transductores de los sistemas de ultrasonido en 20 €.

## 9.2. Hardware de Radiofrecuencia

El hardware de radio se emplea tanto para comunicación como para sincronización. Inicialmente en el prototipo original se ha empleado el módulo nRF2401. Este módulo permite mantener el coste extremadamente bajo dado que tiene un coste de 0,60 € por unidad. Es un módulo complejo que implementa un protocolo de envío múltiple con gestión de acceso al medio, control de errores y reenvío automático. Es capaz de transmitir a una velocidad de 2 Mbps con un alcance de hasta 100 metros en espacios abiertos, siendo capaz de ampliar esta cobertura hasta 1 Km incorporando una antena externa y un amplificador.

Inicialmente se pretendía emplear el nRF2401 tanto como sistema de sincronización como sistema de comunicación. Este dispositivo permite configurar decenas de parámetros para controlar el flujo, errores, paquetes de ACK (ACKnowledgement), potencia de emisión, acceso al medio y reintentos de envío. En las pruebas iniciales se detectaban grandes variaciones en cada una de las medidas que excedían con mucho los márgenes de error de los transductores.

Después de estudiar los motivos de estas variaciones, se detectó que la pila de protocolos del nRF2401 ocasionaba estos retardos. El procesamiento de la información de los paquetes y la puesta a disposición en las capas superiores de la información de recepción de los paquetes originaba que el instante de disponibilidad de la información de que había llegado el paquete de radio fuera muy variable.

Después de realizar cambios en la configuración del módulo para reducir las posibles variaciones de tiempo como eliminar los reenvíos, eliminar las esperas por paquetes de ACK y eliminar la espera aleatoria por el acceso al medio, se habían reducido las variaciones y los errores de las medidas, pero no se han conseguido eliminar, con lo que ha sido necesario buscar otras alternativas y separar el hardware de comunicaciones empleado para la comunicación y el empleado para sincronización.

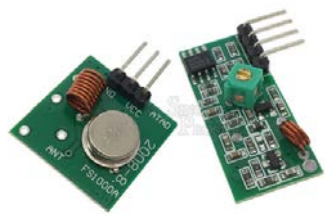

*Figura 16.Imagen de un transceptor de radiofrecuencia de 433 MHz.*

Para intentar eliminar las variaciones aleatorias producidas por la gestión de la pila de protocolo del módulo nRF2401 ha sido necesario buscar un hardware de sincronización mucho más simple, que no introdujera retardos debido a la gestión interna de los protocolos. Se ha seleccionado el hardware de emisión de radio más simple del que se puede disponer. En la Figura 16 puede verse el módulo emisor y el módulo receptor de modo independiente. El transmisor únicamente permite emitir una portadora de 433 MHz. Por otra parte, el receptor detecta la portadora y pone a nivel alto la patilla de salida. Como puede verse, entre los dos simulan un canal de comunicación de un solo bit con errores, dado que constantemente este bit está fluctuando de 0 a 1 debido a las interferencias, pero al no implementar ninguna lógica no puede incorporar retardos por el procesamiento de la lógica interna.

El módulo de transmisión emplea modulación ASK (*Amplitude Shift Keying*).

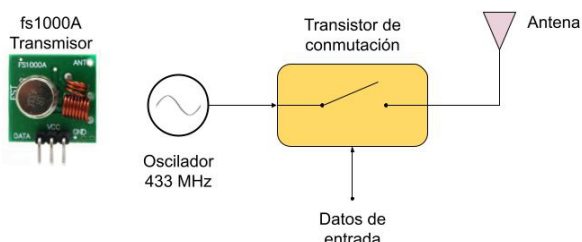

*Figura 17.Esquema básico de modulación ASK.*

Su hardware es muy sencillo, cuando se aplica el valor *ALTO* a la entrada se genera la onda de radio y cuando se aplica el valor *BAJO* deja de transmitir. El cambio de estado la realiza el transistor de conmutación (ver Figura 17).

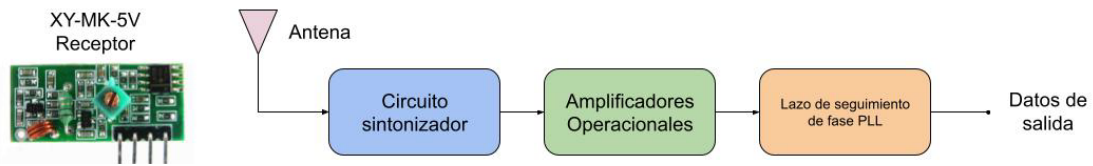
*Figura 18. Esquema del circuito receptor*

El módulo receptor simplemente se sincroniza con la fase de la onda de radio con el circuito sintonizador, la amplifica y entra en el lazo de seguimiento de fase (ver Figura 18). Las pruebas han demostrado que la lectura no es consistente en el tiempo, perdiendo la fase a los pocos milisegundos, por lo que es necesario realizar cambios en los datos de entrada de forma constante.

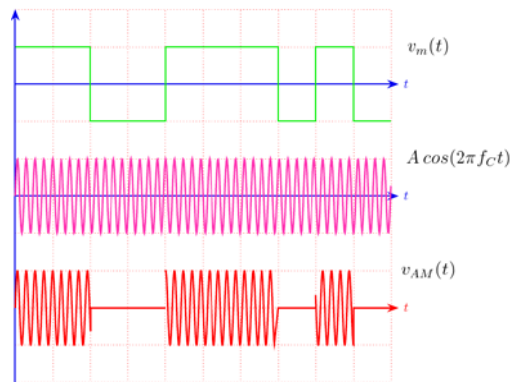
*Figura 19. Modulación de la señal de los módulos SRF04.*

En la Figura 19 puede verse como se combina la entrada de datos (en verde) con la salida de la onda del oscilador (en lila) para generar la onda de radio que se transmite al aire (en rojo).

### 9.3. Comunicación y sincronización

En el protocolo se envía una comunicación inicial desde el robot para comunicación y sincronización. En el paquete inicial se envía la información del código del robot que solicita la operación de localización. Esta información se transmite codificada mediante el hardware de sincronización. Este mismo paquete se emplea como sincronización para iniciar el temporizador de contabilización de tiempo de vuelo de la señal de ultrasonidos.

Al finalizar la operación de localización, cada una de las balizas debe enviar la información del tiempo de vuelo al robot para que calcule su posición. En este caso, varias balizas pueden transmitir a la vez su información, por lo para estas transmisiones se emplean los transceptores nRF2401 dado que gestionan el acceso al medio concurrente y el control de errores.

### 9.4. Hardware de temperatura y medidas inerciales

La temperatura y humedad son dos factores que inciden en la velocidad de transmisión de las ondas de ultrasonidos en el medio, por este motivo se ha incorporado un sensor inercial CMPS09. Este sensor tiene un giróscopo, un acelerómetro y un magnetómetro junto con un sensor de temperatura.

### 9.5. Hardware de procesamiento de señales (Microcontrolador)

Para controlar de todos los sensores, en las balizas se empleará un microcontrolador atmega328 (Figura 20) de la marca ATMEL. Este es uno de los microcontroladores más difundidos en la actualidad y forma parte de una de las placas de desarrollo más conocidas mundialmente, el Arduino Uno R3.

El atmega328 se presenta en un encapsulado de 28 pines. Internamente tiene un procesador de 8 MHz con 32 KB de Flash, 2KB de RAM y 1KB de EEPROM. El microcontrolador tiene un reloj interno, por lo que solo es necesario conectar la alimentación para activarlo.

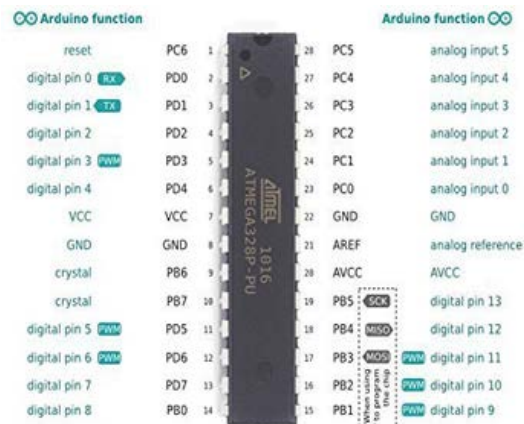

*Figura 20.Encapsulado ATMEga328.*

## 10. Composición de las balizas

Cada baliza está compuesta por el subsistema de alimentación, el subsistema de control, el subsistema de comunicación de radio y el subsistema de recepción de ultrasonidos (Figura 21).

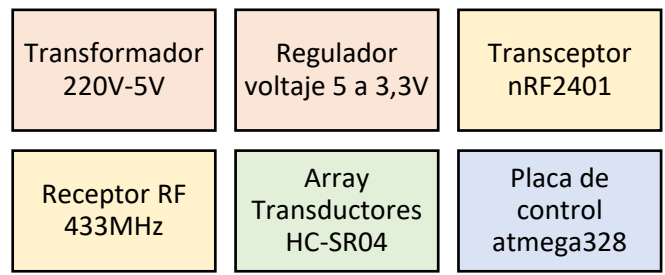

*Figura 21.Componentes modulares de las balizas.*

### 10.1. Descripción de los módulos

La baliza se compone de los siguientes módulos:

- **Transformador de 200V-5V**: el hardware de consumo normal tiene un microtransformador con salida de 5 V.
- **Regulador de 5V a 3,3V**: la placa de la baliza posee un transceptor nRF24L01 que funciona a 3,3V, por lo que es necesario reducir la tensión del transformador principal de 5V a 3,3V.
- **Transceptor nRF24L01**: la baliza posee un transceptor de radio con capacidad de procesamiento desatendido de la pila de protocolos de comunicación, con identificación de errores mediante códigos CRC, con opción de envío confirmado con paquetes de ACK y gestión de múltiples canales.

- **Receptor de radio de 433 MHz:** es un módulo simple de envío de información con portadora OOK de 433 MHz. Se emplea como sistema de sincronización para las medidas de los tiempos de vuelo, dado que no tiene retardos por la gestión de los protocolos de comunicación.
- **Array de transductores de ultrasonidos:** cada baliza contiene un array de transductores compuestos por módulos HC-SR04 modificados para operar solo como receptores. Estos módulos identifican la recepción del paquete de pulsos de ultrasonidos. Con una array de 4 transductores, las balizas son capaces de cubrir 90 grados de arco.
- **Placa de control Atmega328:** se ha diseñado una placa específica para dar soporte a un ATmega328 como microcontrolador principal. Este microcontrolador controla todos los sensores y el hardware de comunicaciones.

## 10.2. Diseño de la estructura

Para empaquetar todos los componentes que forman parte de la baliza ha sido necesario diseñar una caja con la forma apropiada para que los sensores de ultrasonidos cubran el arco de 90°. En la Figura 22 puede verse esta caja en la versión de bajo consumo.

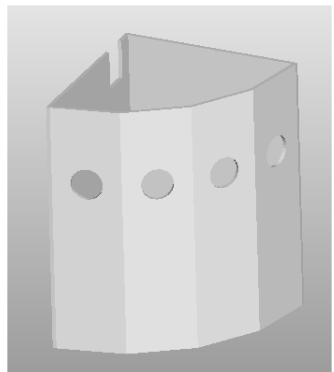

*Figura 22.Diseño 3D de la caja de las balizas.*

Esta versión necesita bastante espacio para incorporar dos packs de pilas para proporcionar de modo independiente el voltaje necesario para alimentar los componentes que trabajan a 3,3 y 5 voltios. En la versión alimentada por corriente alterna a 220 V el tamaño de la caja puede reducirse a la mitad.

## 10.3. Diseño del hardware de consumo estándar

Ha sido necesario diseñar la placa de circuito impreso (ver Figura 23) para montar los componentes de control. Esta placa tiene versiones alimentadas con 220 V y la versión alimentada por baterías. La placa está diseñada para poder ser empleada en un sistema de control domótico dentro de la casa, pudiendo conectar múltiples tipos de sensores y actuadores. Las placas poseen un módulo de corte de alimentación de los sensores y actuadores para reducir la energía consumida. Adicionalmente todas las placas poseen un módulo de comunicación de radio nRF2401 para comunicarse entre ellas y con el módulo de control central. El módulo de control está basado en el microcontrolador atmega328 que necesita una tensión de alimentación entre 3 y 19 voltios para lo que se ha integrado en la placa un módulo compacto de transformación de 220V a 5V.

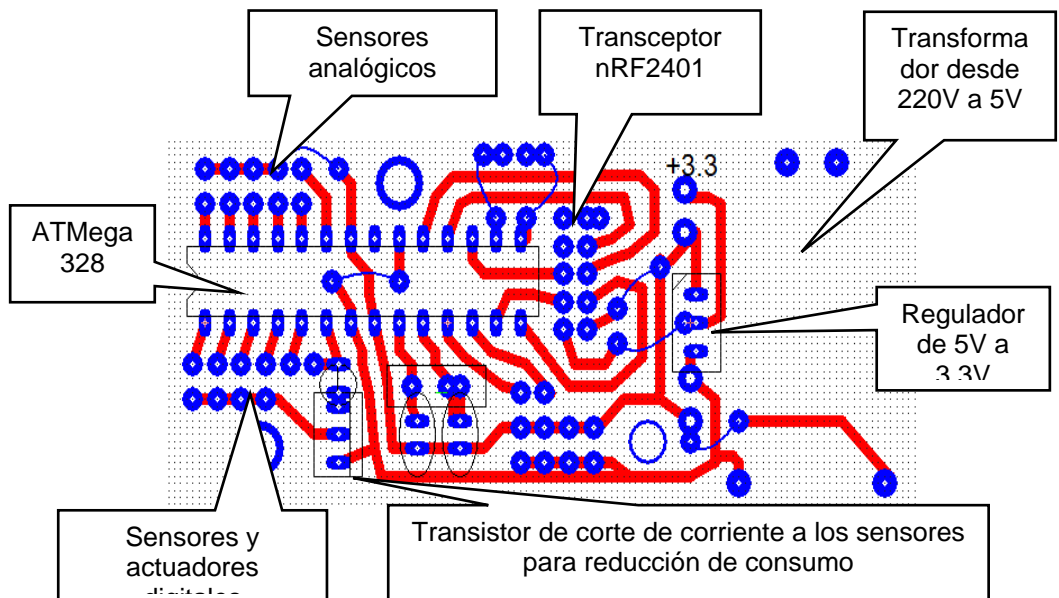
*Figura 23.Placa PCB de la baliza de conexión directa a corriente de 220V.*

En la siguiente Figura 24 puede verse una imagen de la placa de control de las balizas con el microtransformador de 200V a 5V.

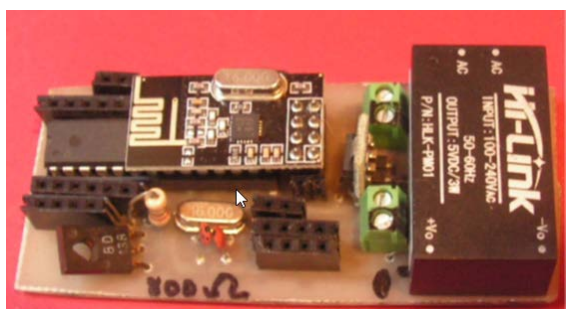
*Figura 24.Baliza para conexión directa a 220V AC.*

En la Figura 24 se muestra la misma baliza, identificando todos sus componentes.

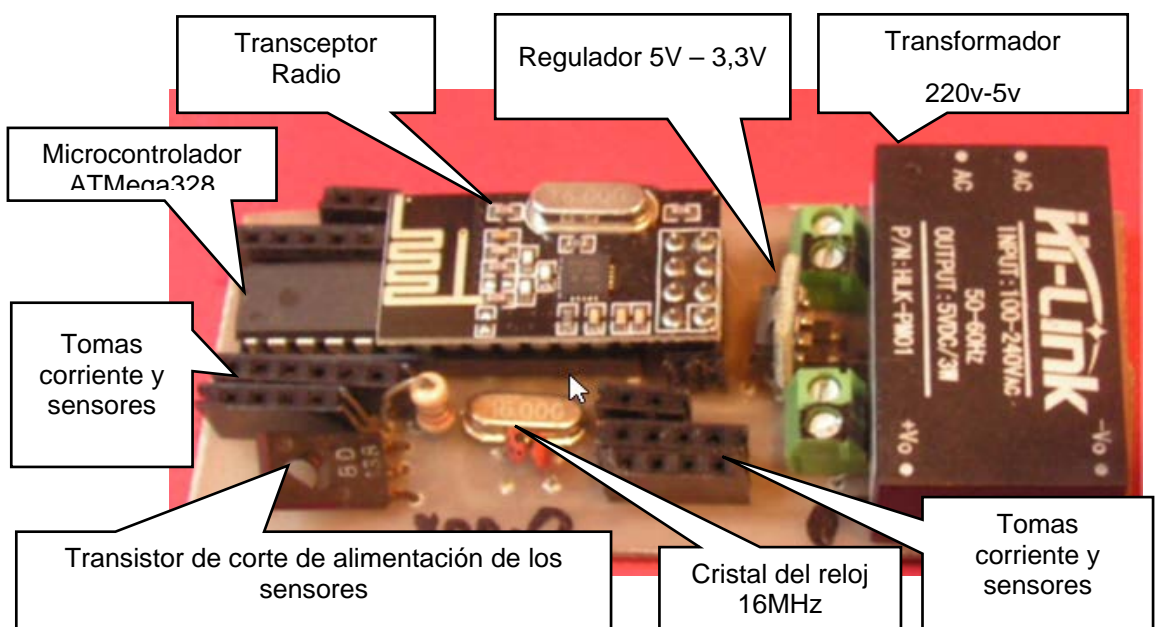
*Figura 25.Módulo baliza con fuente de alimentación.*

La placa empleada para las balizas de localización es exactamente la misma que la mostrada en la Figura 25, añadiendo el módulo de radio simple de transmisión de portadora a 433MHz.

### 10.4. Diseño del hardware de bajo consumo

Dado que SAPPO está diseñado para operar en viviendas estándar no cableadas, en algunos lugares donde es necesario situar una baliza para conseguir ampliar al máximo su cobertura, puede ser posible que no sea fácil llevar un cable de alimentación, o que directamente no quiera llevarse por cuestiones estéticas. En este caso hay que emplear balizas alimentadas por baterías. Para estos casos se ha diseñado una placa especial sin reguladores de tensión y con transistores de corte de alimentación de corriente a los sensores que se alimenta directamente con dos packs de baterías, uno para proporcionar 3,3 V y otro para proporcionar los 5 V que necesitan el resto de los sensores.

A pesar de no emplear reguladores, el atmega328 sigue teniendo un consumo bastante elevado como para poder estar alimentado por baterías durante largos periodos de tiempo, por lo que es necesario emplear los modos de bajo consumo.

El módulo de bajo consumo se basa en el desarrollo de la placa anterior permitiendo la conexión directa de packs de baterías de 5V y de 3,3V (ver Figura 26), dado que el transceptor de radio frecuencia opera a un voltaje de 3,3V, mientras el resto de los componentes y sensores operan a 5V.

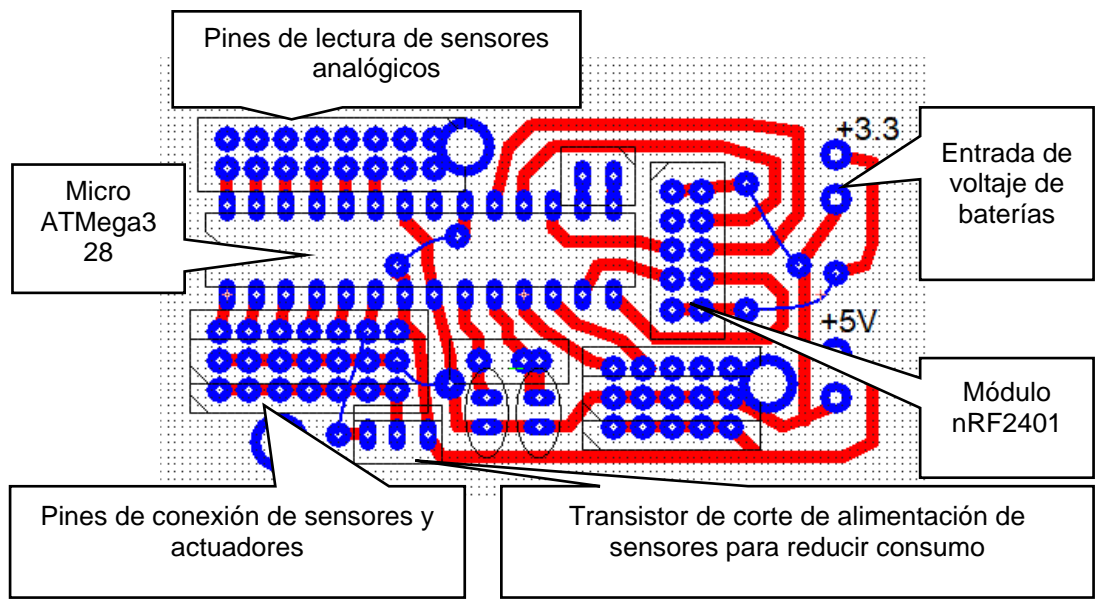

*Figura 26. PCB de baliza de alimentación por baterías.*

La capacidad de reducción de consumo viene implementada en el firmware del microcontrolador. Este micro posee varios modos de sueño en los que reduce el consumo apagando distintos dispositivos o subsistemas:

- **Modo *Idle*:** este modo detiene únicamente la CPU, el control de interrupciones y los relojes, el conversor analógico digital, etc. siguen funcionando, con lo que no se ven afectados los dispositivos. Este es un modo básico de suspensión.
- **Modo *ADC noise reduction*:** este modo detiene los relojes I/O, CPU y FLASH. Este modo no altera las medidas del conversor analógico-digital permitiendo mantener la resolución de medidas. Dado que siguen activos gran parte de los sistemas, se puede despertar de este modo de trabajo mediante un reset externo, una interrupción del Watchdog, un reset del Brown-out que controla el nivel de alimentación, una interrupción del Timer/Counter2 o una interrupción externa en INT0 o INT 1 entre otras.
- **Modo *Power-down*:** en este modo de trabajo el oscilador externo se detiene. Las interrupciones externas y el Watchdog continúan operando si están habilitados. En este modo se paran todos los relojes, solo es posible despertar con una interrupción asíncrona externa o con una interrupción del Watchdog. Es uno de los modos más empleados para conseguir un alto ahorro de energía. En caso de no disponer de un dispositivo que genere una interrupción externa, es habitual generar múltiples periodos de letargo seguidos de cortos periodos de actividad en los que se comprueba el estado para verificar si se sigue en estado activo o se vuelve a un periodo de letargo.
- **Modo *Power-save*:** es similar al modo anterior, pero se mantiene activo el Timer/Counter2. Permite emplear este timer para despertar, a costa de un mayor consumo de energía. Si no es necesario, mejor emplear el modo anterior.
- **Modo *Stand-by*:** su comportamiento igual a Power-down con la diferencia de que el oscilador sigue en funcionamiento. Emplea 6 ciclos para activarse.

### .10.4.1. Comparativa de consumo con bajo consumo

En la siguiente tabla se ven los datos comparativos de consumo empleando los distintos modos de sueño del microcontrolador ATMega328. La electrónica de las balizas está diseñada para operar en Power-down con el ATmega328 estándar con cristal externo de 16 MHz, lo que ofrece un consumo de 0,03 mA como se puede ver en la Tabla 1.Consumo de los modos de sueño del chip Atmega328.. Esto implica que, si instalamos una batería de 2000 mAh, en modo de espera tendría una duración de más de 7 años que se vería reducida por el tiempo de actividad debido a las solicitudes de posicionamiento de los robots.

*Tabla 1.Consumo de los modos de sueño del chip Atmega328.*

| Modo de sueño | 16 MHz | 8 MHz |
|---|---|---|
| Power-down (placa Arduino) | 0,15 mA | - |
| Power-down(cristal externo) | 0,03 mA | 0,03 mA |
| Idle | 12 mA | 10,5 mA |
| ADC Noise reduction | 10,9 mA | 8 mA |
| Power-save | 2,9 mA | 1,3 mA |
| Stand-by | 1,3 mA | 0,4 mA |

## 10.5. Modificación del algoritmo base para adaptarlo a las balizas de bajo consumo

En algunas ocasiones, el lugar más favorable para conseguir la mejor cobertura puede que no esté situado cerca de ninguna toma de electricidad y puede que sea complicado cablear la alimentación hasta la baliza. Para estos casos, se han diseñado las balizas de bajo consumo, tanto para una instalación temporal como definitiva. Estas balizas se alimentan con dos packs de baterías para garantizar la descarga homogénea de todos los módulos de carga. Uno de los packs alimenta los dispositivos de 5V y otro los de 3,3 V. No emplea reguladores de tensión para evitar el consumo ocasionado por esta regulación. En el hardware se implementa un transistor de corte de corriente a los sensores y actuadores para garantizar que cuando el microcontrolador se encuentre en modo sueño, ninguno de los dispositivos se encuentre consumiendo energía.

En el caso de que en una habitación exista una o varias balizas de bajo consumo, estas deben ser despertadas en cuanto el robot entre en la habitación, para ello, en cuanto pierda el enlace con las balizas de la habitación de la anterior habitación o ya desde la anterior habitación si no queremos tener retardos, el robot, que conoce los códigos y el tipo de balizas existentes en cada habitación enviará de modo indefinido durante 5 segundos paquetes para solicitar que despierten las balizas, indicando los códigos de baliza que quiere despertar para evitar que despierten otras balizas en el radio de cobertura del paquete de RF.

El microcontrolador de cada baliza de bajo consumo entrará en modo *Power-down* por un periodo de 5 segundos. Cada 5 segundos se despertará, activará solo el módulo de comunicaciones nRF2401 durante 100 ms. Si durante este tiempo no se detecta la recepción de un paquete en el que se solicite que despierte, volverá a modo sueño. En el caso de que detecte el paquete se activará y esperará la solicitud de medida.

Volverá a pasar a modo sueño si el robot indica mediante un paquete de RF que no necesita más lecturas por haberse parado o si transcurre más de 1 minuto sin que se le solicite una lectura.

Con este modo de comportamiento generamos un pequeño retraso en la activación que puede ser eliminado despertando la baliza cuando el robot se encuentra todavía en la habitación anterior. Como contrapartida generamos una baliza que puede mantenerse en espera más de 4 meses con un pack de 5 pilas de formato de tamaño AA.

## 11. Algoritmo de obtención de medidas concurrentes

El algoritmo básico de recuperación de una lectura involucra las comunicaciones de radio y ultrasonidos necesarias para recuperar el tiempo de vuelo desde el robot móvil a todas las balizas y la comunicación de vuelta de las medidas desde las balizas hasta el robot móvil. Si contamos con un solo robot, no es necesario complicarlo más, pero en el caso de que exista la posibilidad de que dos o más robots soliciten sus medidas de modo concurrente, es necesario gestionar el acceso al medio ultrasónico, dado que al emplear ultrasonidos de banda estrecha no es posible la emisión simultánea de varios robots. En los siguientes apartados se describirán los tipos de paquetes necesarios y el papel que juegan dentro del algoritmo de comunicaciones, para gestionar un sistema de acceso al medio del tipo CSMA/CA mediante colas circulares con paso de testigo.

### 11.1. Algoritmo de lectura simple, lecturas múltiples y señales reflejadas

En el esquema de la Figura 27 se visualizan las comunicaciones que se realizan entre el robot y las balizas para obtener una localización. Una vez que el robot inicia la solicitud de la operación de medida con el envío del paquete de radio de sincronización, se envía un lote de pulsos de ultrasonidos.

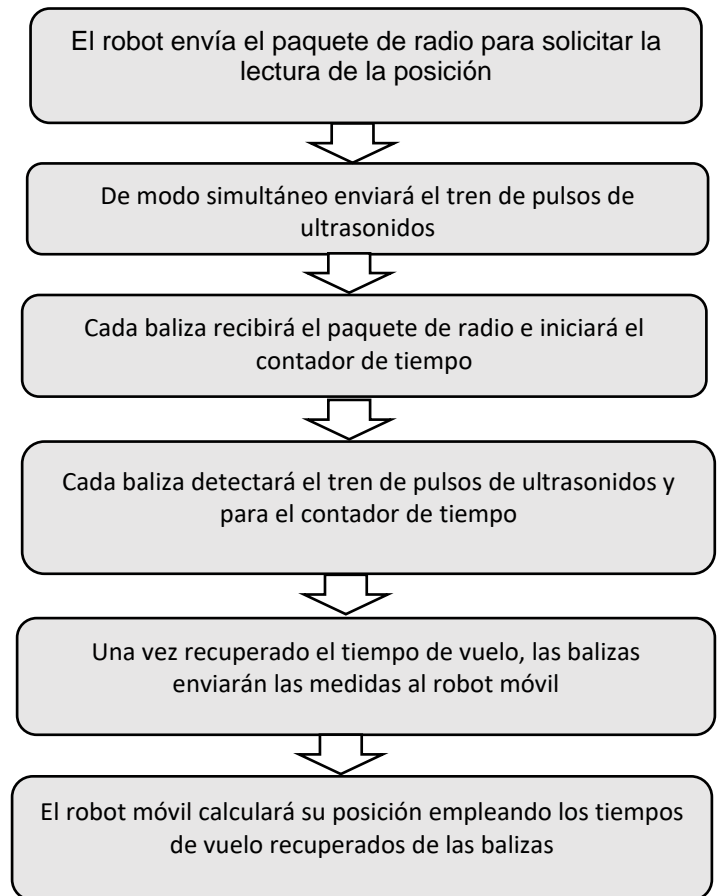
*Figura 27.Pasos para efectuar una lectura.*

El robot espera la recepción de la confirmación de llegada de todas las balizas. En este punto tenemos dos algoritmos diferentes para aplicar.

### .11.1.1. *Algoritmo de espera por la atenuación del pulso*

En uno de ellos, simplemente enviamos el tren de pulsos de ultrasonidos y esperamos hasta que la energía del pulso se disipe. Teniendo en cuenta un alcance de 9 metros y conociendo la velocidad de desplazamiento de pulso de ultrasonidos, podemos calcular con precisión el tiempo que tardará en disiparte la energía del pulso de ultrasonidos, de forma que no pueda interferir con el siguiente pulso de ultrasonidos enviado. Este procedimiento es muy sencillo de implementar, pero limita la frecuencia de las lecturas de posición, dado que, en la mayoría de los casos, las balizas se encontrarán a una distancia inferior al máximo teórico.

### .11.1.2. *Algoritmo de espera por la confirmación de las balizas*

Por este motivo, el algoritmo que se implementa en SAPPO pretende agilizar e incrementar el número de lecturas por segundo, para ello, por cada una de las balizas, en cuanto detecten el pulso de ultrasonidos, se procederá al envío de un paquete de confirmación de recepción. Una vez que el robot recupere todos los paquetes de confirmación de recepción, el robot envía el siguiente pulso de ultrasonidos.

Este sistema agiliza e incrementa la frecuencia de lecturas por segundo, pero genera una serie de efectos secundarios de difícil solución. Dado que podemos tener dos trenes de pulsos de ultrasonidos en el aire con energía suficiente para que sea posible su detección, las balizas pueden detectar el primer pulso procedente de alguna trayectoria con rebotes, una vez iniciado el siguiente ciclo de detección de lectura. Esto originaría que recupere una medida de distancia totalmente incorrecta. Por consiguiente, para acelerar la frecuencia de lecturas de distancia, será necesario diseñar un algoritmo que sea capaz de diferencias las lecturas directas de las lecturas de señal con rebotes.

Dada la construcción de las balizas, resulta relativamente sencillo identificar el ángulo de incidencia de la señal, aunque con cierto grado de imprecisión. Debe tenerse en cuenta que cada baliza cuenta con 4 transductores para cubrir 90 grados. Cada uno de ellos cubre un ángulo de 22,25 grados. Para conseguir corregir las medidas procedentes de rebotes se dispone de dos técnicas.

- Por un lado, pueden identificarse las medidas cuya varianza excede de cierto valor, siempre pudiendo confirmar estos límites con los datos de velocidad y orientación del robot. Debe tenerse en cuenta que la variación de la distancia obtenida de cada baliza debe ser consecuente con la velocidad de desplazamiento del robot.
- Adicionalmente se puede intentar diferenciar los ángulos de incidencia de las señales directas y reflejadas, dado que, si los ángulos son suficientemente diferentes, serán detectados por transductores distintos en la baliza. Si se consigue este objetivo, pueden detectarse señales reflejadas identificando el ángulo de incidencia de la señal reflejada

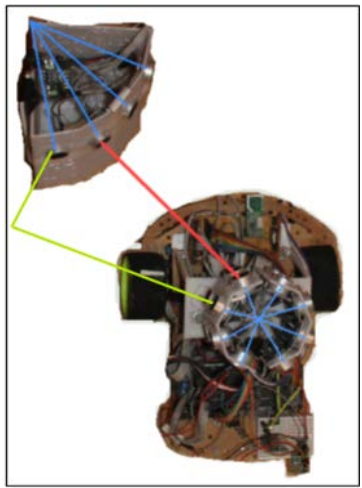

*Figura 28.Multirruta*

Como se puede ver en la Figura 28, en caso de que el emisor emita por todos los transductores a la vez, si existe una pared al lado del robot, la señal de ultrasonidos que choca contra la pared saldrá rebotada en un ángulo similar a la señal real cuyo tiempo de vuelo es menor. En este caso sería muy complicado detectar o diferenciar la señal reflejada de la principal, si bien, hay que tener en cuenta que la diferencia de la longitud de ambas trayectorias hará que los tiempos de vuelo sean muy similares, del orden de microsegundos, lo que propiciará que sea imposible solicitar una nueva operación de lectura en esos tiempos tan cortos.

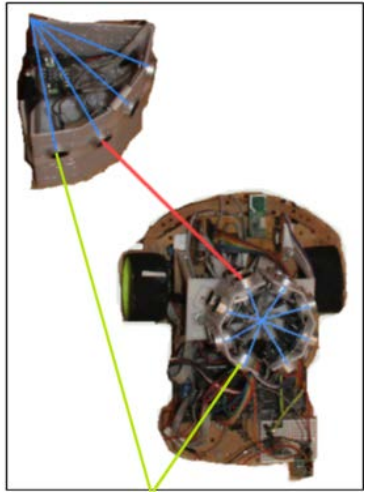
*Figura 29. Multirrutas opuestas.*

En el caso mostrado en la Figura 29 se puede ver un emisor que emite por todos los transductores a la vez, en una habitación típica con paredes contrapuestas y las balizas situadas en una de ellas. La señal de ultrasonidos que choca contra la pared saldrá rebotada en un ángulo similar a la señal real cuyo tiempo de vuelo es menor. En caso de que fuera posible iniciar la segunda lectura, sería muy complicado detectar o diferenciar la señal reflejada de la principal. Para que el control por varianza funcione, la pared deberá encontrarse en un rango de distancias muy concreto, dado que, en cualquier otro caso, la señal reflejada podría recuperarse como una señal real del siguiente ciclo de lectura. Este problema podría solucionarse si no se emite radiación por los transductores opuestos a las balizas. Esta actuación resulta compleja y es totalmente dinámica, con lo que es necesario controlar en todo momento la orientación del robot con respecto a las balizas. Igualmente es muy dependiente de la colocación de las balizas, dado que, si se encuentran en puntos opuestos, no sería posible emplear esta técnica y sería necesario emplear una técnica dinámica de cálculo de probabilidad de rebotes y trayectorias posibles de los mismos para identificar las posibles señales erróneas.

## 11.2. Resolución del posicionamiento por trilateración

Para la obtención de la localización en el espacio es necesario tener un mínimo de tres balizas. En la Figura 30 se ve una representación de la intersección de las tres balizas. Normalmente en los sistemas de posicionamiento no es normal encontrar balizas con emisión omnidireccional, de forma que cubran el volumen total de una circunferencia, por lo que el volumen cubierto dependerá de la tecnología empleada. En el caso de SAPPO, las balizas normales con 4 transductores cubren 90 grados de arco en horizontal y en vertical, el volumen depende de la apertura del cono de radiación de 30 grados, lo que implica que el volumen crece con la distancia del punto de posicionamiento a la baliza.

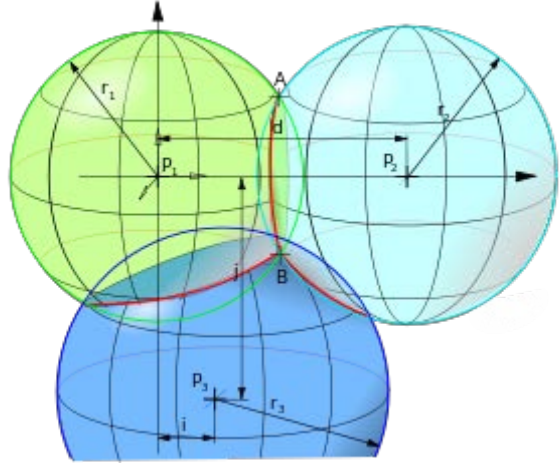
*Figura 30. Intersección esférica.*

Con las Ecuaciones (4) a (10) se calculan los puntos de intersección de las tres circunferencias. Mediante trilateración esférica es posible obtener el punto de intersección de las tres esferas que pueden dar como resultado 0, 1 o dos puntos de intersección. En el caso común de obtener dos puntos necesitamos discriminar uno de ellos, o empleando una cuarta medida, o bien porque se encuentra en una ubicación inconsistente con el mapa de navegación del robot.

$$r_1^2 = x^2 + y^2 + z^2 \tag{4}$$

$$r_2^2 = (x-d)^2 + y^2 + z^2 \tag{5}$$

$$r_3^2 = (x-i)^2 + (y-j)^2 + z^2 \tag{6}$$

$$x = \frac{r_1^2 - r_2^2 + d^2}{2d} \tag{7}$$

$$y^2 + z^2 = r_1^2 - \frac{(r_2^1 - r_2^2 + d^2)^2}{4d^2} \tag{8}$$

$$y = \frac{r_1^2 - r_3^2 - x^2 + (x-i)^2 + j^2}{2j} = \frac{r_1^2 - r_3^2 + i^2 + j^2}{2j} - \frac{i}{j}x \tag{9}$$

$$z = \sqrt{r_1^2 - x^2 - y^2} \tag{10}$$

En SAPPO, dado que la navegación se realizará en el plano, aunque exista una distancia entre la altura de las balizas y la altura del emisor del robot, esta es conocida en cada una de las habitaciones, por lo que es posible reajustar las distancias obtenidas para calcular la posición empleando sólo la intersección de dos circunferencias.

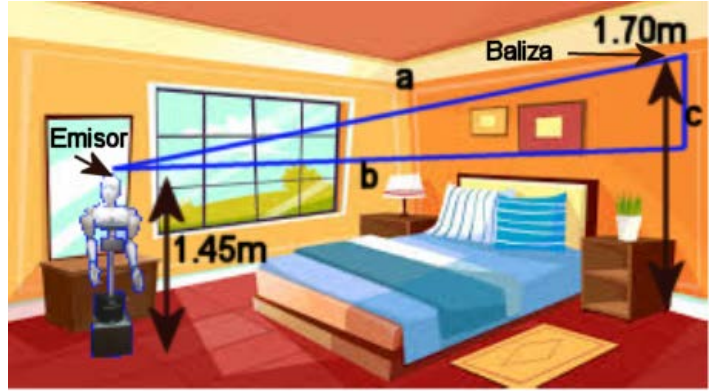

*Figura 31.Reajuste de las medidas.*

En la Figura 31 puede observarse la localización de la baliza, a 1,70 m del suelo y la localización del emisor del sistema de localización situado en la cabeza del robot. Inicialmente no puede emplearse el posicionamiento en el plano, dado que no comparten plano, pero dado que se conoce la altura de la baliza y la altura del robot, solo hay que resolver el segmento $b$ del triángulo rectángulo mediante las Ecuaciones (11) a (14).

$$a^2 = b^2 + c^2 \qquad (11)$$

siendo,

$$c = 1{,}70 - 1{,}45 \qquad (12)$$

$$a = resultado\ de\ la\ medida = medida\_distancia \qquad (13)$$

$$b = \sqrt{medida\_distancia^2 - 0{,}25^2} \qquad (14)$$

de forma que para el cálculo de los radios de las circunferencias de intersección se emplearán los valores de $b$, en vez de las medidas reales de distancia.

Con este sencillo cambio, las fórmulas simplificadas corresponden a las Ecuaciones (15) a (19).

Comenzamos por la representación estándar de las circunferencias,

$$(x - x_1)^2 - (y - y_1)^2 = r_1^2 \qquad (15)$$

$$(x - x_2)^2 - (y - y_2)^2 = r_2^2 \qquad (16)$$

desarrollando los binomios,

$$(x^2 - 2 \cdot x \cdot x_1 + x_1^2) - (y^2 - 2 \cdot y \cdot y_1 + y_1^2) = r_1^2 \qquad (17)$$

$$(x^2 - 2 \cdot x \cdot x_2 + x_2^2) - (y^2 - 2 \cdot y \cdot y_2 + y_2^2) = r_2^2 \qquad (18)$$

y restando las dos ecuaciones,

$$(2x(x_2 - x_1) + x_1^2 - x_2^2) - (2y(y_1 - y_2) - y_1^2 + y_1^2) = r_1^2 - r_2^2 \qquad (19)$$

se obtiene la ecuación de la recta que pasa por los dos puntos de intersección, con lo que es posible despejar una de las incógnitas dado que los valores de $x_1, x_2, y_1, y_2, r_1$ y $r_2$ son conocidos. Una vez despejada debe ser sustituida en la Ecuación (17) para resolver la otra incógnita y dado que el resultado de despejar la Ecuación (17) será una ecuación de segundo grado, pueden obtenerse hasta dos resultados válidos.

## 12. Problemática de precisión en sistemas de localización por ultrasonidos con arrays de transductores como SAPPO

La cobertura omnidireccional que puede proporcionar el array de transductores genera un efecto indeseado al disponer de múltiples transductores ubicados en lugares diferentes. Dado que los tiempos de vuelo se calculan desde el transductor más cercano de la estación móvil a la baliza, se generan cambios bruscos de distancia cuando debido a la orientación entre ambos, los transductores más cercanos cambian y pasan a ser otros. En estático, el cálculo de distancia genera medidas con una precisión muy estable, dado que corresponden con las distancias entre el transductor del robot y el transductor de la baliza que se encuentran más próximos (ver Figura 32) y estos no varían porque el robot no se mueve. En cuanto cambia la orientación entre el anillo emisor y las balizas, el camino más corto puede involucrar nuevos transductores, lo que modifica la geometría de las medidas.

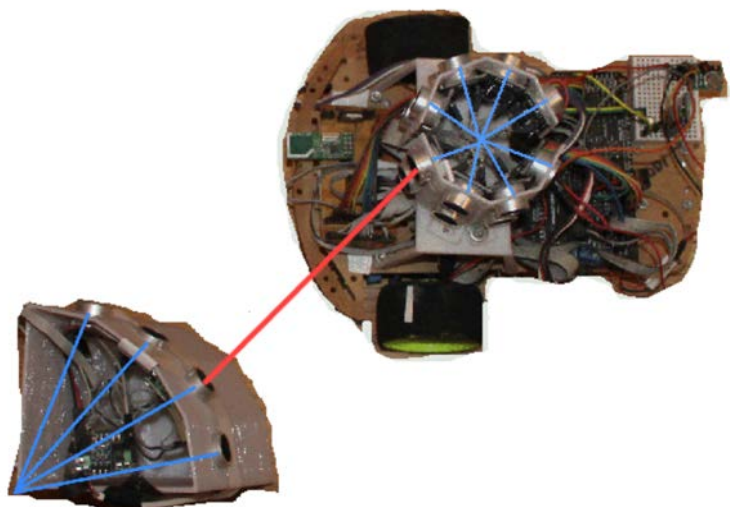

*Figura 32.Puntos centrales de referencia para cálculo de distancia y recorrido del ultrasonido.*

Para intentar aproximar las medidas del modo más preciso posible, minimizando las diferencias de distancia originadas por el salto de transductores, debe calcularse en todo momento la distancia entre el centro del polígono de cada baliza y el centro del polígono del anillo emisor del robot.

Debe tenerse en cuenta que tanto el anillo de transductores del robot como las balizas tienen forma de polígono. En el caso del robot tenemos dispositivos con polígonos de ocho y doce lados (para RODI ocho y para DANI doce) y en el caso de las balizas, lo más común es que estén compuestos por la cuarta parte de un polígono de ocho lados.

Para calcular la distancia entre el punto central del anillo del robot y el punto central del anillo de la baliza debemos calcular las dos apotemas (Figura 33), la correspondiente al robot y a la baliza.

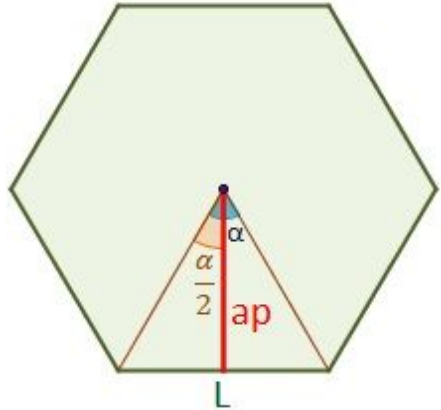

*Figura 33.Cálculo de la apotema (punto central de posicionamiento).*

En cada polígono conocemos la longitud de su lado. En la Figura 34 se identifican los parámetros necesarios para realizar el cálculo de la apotema empleando las medidas de la longitud de su lado.

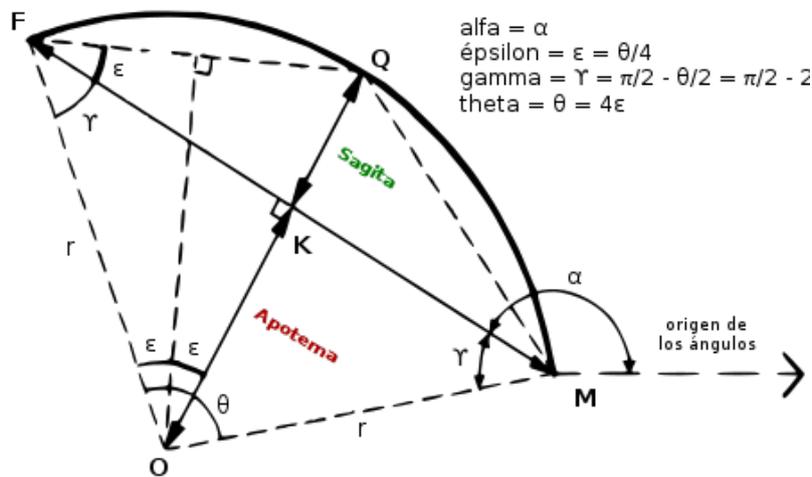

*Figura 34.Cálculo de la apotema .*

Con la Ecuación (20) se calcula la apotema, para lo que se necesita obtener los valores tanto del radio como del lado del polígono.

$$a = \sqrt{r^2 - \left(\frac{l}{2}\right)^2} \qquad (20)$$

Como el radio no es conocido de entrada, de la Ecuación (5.21) obtenemos el radio del polígono a partir de su lado

$$r = \left(\frac{l}{2 \cdot sen\left(\frac{360°}{2 \cdot n}\right)}\right) \qquad (21)$$

En la Figura 35 podemos ver el polígono que forma el anillo emisor del robot móvil y se marcan en rojo tanto las apotemas como la distancia máxima hasta el centro. Los transductores se ubican en la mitad de cada uno de los lados, por lo que la apotema es la distancia desde el centro del polígono a cada uno de los transductores.

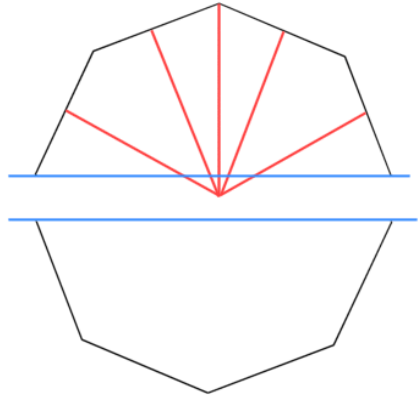

*Figura 35.Geometría del array de emisores con el punto central de localización.*

Como se aprecia en la Figura 35 el cálculo es ligeramente más complicado porque los polígonos se dividen en dos partes y cada una de las partes se separa de la otra por una pequeña distancia que viene determinada por las necesidades de la geometría del robot. Esta distancia debe ser tenida en cuenta y debe incrementarse a la medida de la apotema dado que debe intentarse aproximar al máximo la medida de distancia. Con la medida de la distancia de las dos mitades de los polígonos deben calcularse las distintas medidas de las apotemas. Si se desea obtener la medida con la máxima precisión, siempre debe incrementarse al resultado de la medida de distancia entre transductores de cada una de las balizas, la suma de la distancia de los valores de las dos apotemas que intervienen en la distancia de la medida, la correspondiente al anillo del robot y la correspondiente a la baliza, dado que, en el robot cada apotema se ve incrementada ligeramente y de forma diferente por el espacio entre los dos lados del polígono.

Para calcular las distancias entre cada uno de los transductores al centro podemos resolver el triángulo que se forma con cada una de estas apotemas modificadas. Comenzaremos por la apotema central modificada.

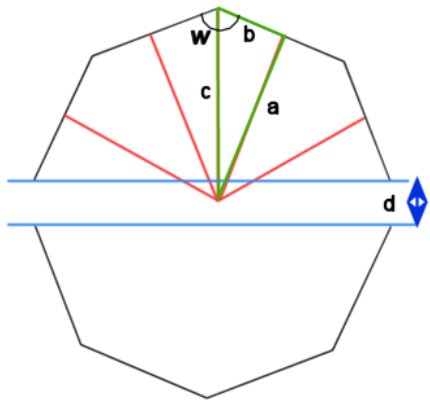

*Figura 36.Cálculo modificado de la apotema.*

En la Figura 36 se puede ver el triángulo que debemos resolver para calcular el valor correcto de la apotema teniendo en cuenta el espacio de separación $d$ entre ambas partes. La apotema modificada central $c$ se calcula como muestra la Ecuación (22).

$$c = r + \frac{d}{2} \qquad (22)$$

y el valor de *r* se obtiene de la Ecuación (5) empleando el valor del lado del polígono *l*, que es un dato conocido. Para continuar se necesita conocer el valor de *b* que es la mitad del lado del polígono (ver Ecuación (23)) y del ángulo w, que dado que el polígono tiene 8 lados es la octava parte de 360 (ver Ecuación (24)). La mitad de *w* se corresponde con el ángulo que forman los segmentos *b* y *c*. Con estos datos puede calcularse la apotema modificada *a* (ver Ecuación (25)) aplicando el teorema del coseno.

$$b = \frac{l}{2} \tag{23}$$

$$w = \frac{360}{8} = 45° \tag{24}$$

$$a = \sqrt{a^2 + c^2 - 2 \cdot a \cdot c \cdot \cos\left(\frac{w}{2}\right)} \tag{25}$$

En la Figura 37 están dibujados todos los parámetros necesarios para calcular el resto de las apotemas modificadas correspondientes a un anillo emisor como el que porta el robot RODI, cuya imagen se muestra en la Figura 37.

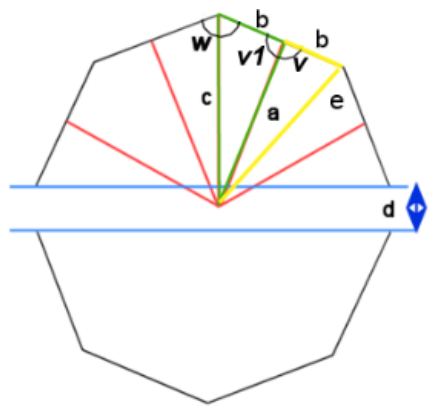

*Figura 37. Radio intermedio.*

Conocidos *a, b, c* es posible obtener todos los ángulos del triángulo *abc* con lo que puede calcularse el ángulo *v1*. Con este valor tenemos *v* con solo restar *v1* de los 180 grados como se puede ver en la Figura 38, con lo que volvemos a tener los dos lados *a, b* y un ángulo *v* pudiendo calcular el lado *e* empleando el teorema del coseno.

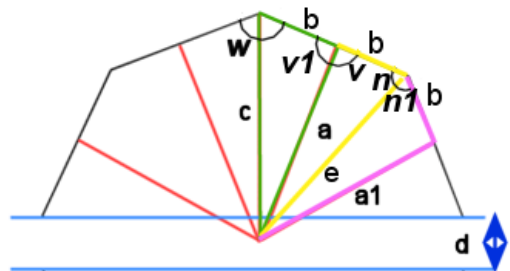

*Figura 38. Segunda apotema modificada.*

En el último paso tenemos de nuevo los lados del triángulo *abe* con lo que podemos calcular el ángulo *n* y con ello obtener el ángulo *n1* restando *n* a la octava parte de 360°, con lo que igualmente volvemos a tener los lados *b, e* y el ángulo *n1* para obtener de nuevo la segunda apotema *a1*.

El resto de las apotemas son una copia reflejada de las ya calculadas. Los tres cuartos de polígono restante tienen las mismas dos apotemas modificadas *a* y *a1*.

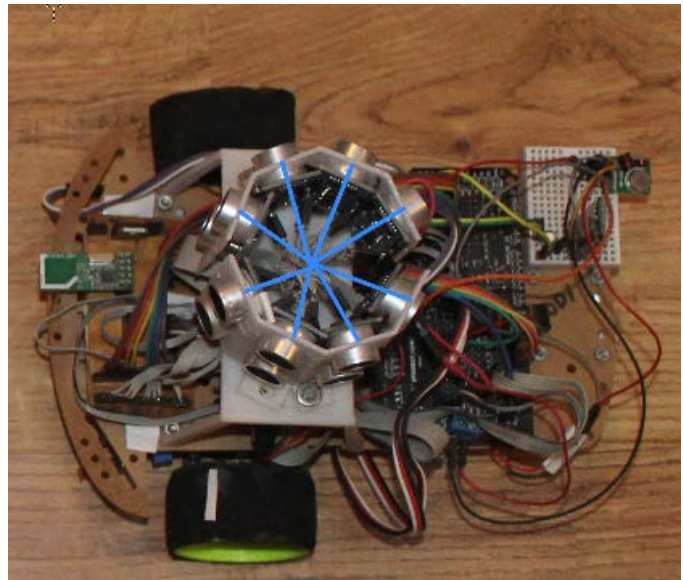

*Figura 39.Vista superior de DANI con el array de sensores y centro de localización.*

En la Figura 40 puede apreciarse de modo muy claro la necesidad de incorporar las apotemas al cálculo de las medidas, pero también puede apreciarse el error inherente a dicha aproximación. Podemos ver marcado en color azul la distancia real recorrida por la onda de ultrasonidos que debe ser medida por el sistema y en amarillo las distancias que debemos calcular y sumar adicionalmente para localizar el punto central del anillo del robot con precisión. Lo deseable es que ambas partes de la medida formaran una recta, pero puede apreciarse que forman un ligero ángulo.

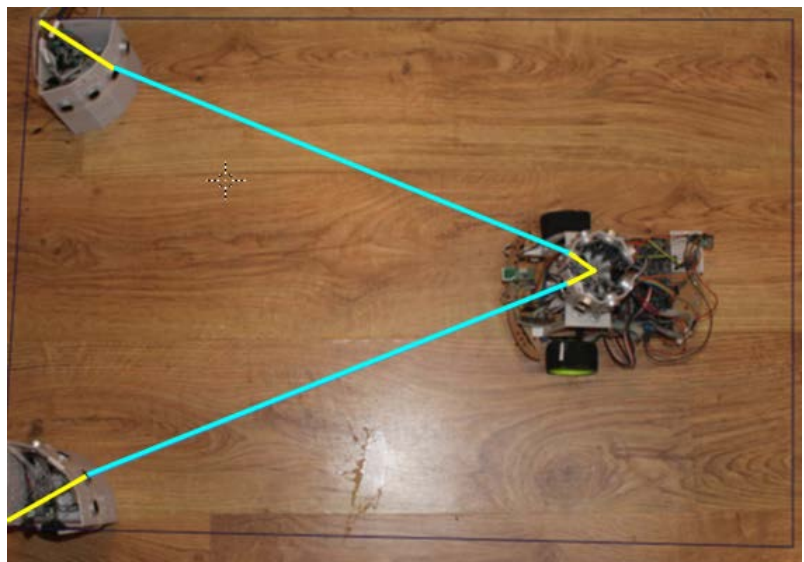

*Figura 40.Distancias totales a dos balizas para calcular la posición del punto central del array de emisores.*

Las apotemas (en color amarillo) siempre parten del centro de los polígonos y terminan en la mitad de sus lados, que es donde se encuentran los transductores, y la distancia recorrida por el tren de pulsos de ultrasonidos (en color azul) siempre parte del centro de los lados de un polígono y es recibido por otro transductor que se encuentra en el centro del lado del polígono de la baliza. El sistema siempre calculará la distancia recorrida entre los dos transductores que se encuentren más próximos. Según estas condiciones, puede apreciarse que en la mayoría de los casos la apotema (en amarillo) formará un ángulo con la trayectoria real de la onda de ultrasonidos, lo que origina un aumento de la distancia medida con respecto a la distancia real de los centros de los polígonos de la baliza y del robot.

Debe tenerse en cuenta que la contribución del error generado por el posicionamiento de los transductores en las balizas y en el anillo emisor, aunque pequeño, es muy variable y afecta de modo distinto dependiendo del posicionamiento angular relativo entre balizas y anillo emisor y también es ligeramente dependiente de la distancia entre anillo y balizas. En cuanto incrementamos la distancia, la contribución del error se reduce, tanto por la reducción de los ángulos relativos entre la trayectoria de los ultrasonidos y las apotemas y por la reducción de la aportación del error en el valor total de la medida.

Debido a la configuración del robot, no siempre la distancia más corta incorpora medidas de los mismos transceptores en el robot móvil y en las balizas. Según se va desplazando el robot, llegará el momento en que nuevos transductores entren en juego para calcular la ruta más corta. Antes de que esto ocurra, la ruta anterior irá acumulando un error en la medida de la distancia real de modo progresivo y lineal que podrá saltar de modo discreto en el momento en que nuevos transceptores entren en la medida.

En la Figura 41 pueden apreciarse dos rutas, una dibujada en color azul y la otra en color verde, de forma que puede verse el máximo punto de giro relativo entre estación móvil y baliza para que los ultrasonidos cambien de la ruta azul a la verde. Puede verse de modo muy claro que la ruta azul de partida llevaba acumulado un error importante con respecto a la distancia real mínima de separación de los dos centros de los polígonos. Se aprecia el punto de partida del centro del anillo del robot móvil, que es el lugar en donde se encontraba el anillo en su posición inicial. En ese punto los transductores implicados en el camino mínimo eran los que formaban parte del camino marcado en azul. Podemos ver la trayectoria de desplazamiento y el punto final en el que se encuentra el anillo. En el punto final las trayectorias más cortas están marcadas por las líneas verdes. Mientras el robot sigue la trayectoria, el error de distancia de la trayectoria azul se ha ido incrementando progresivamente hasta el cambio de trayectoria a la trayectoria verde, momento en el cual se produce un salto de error discreto.

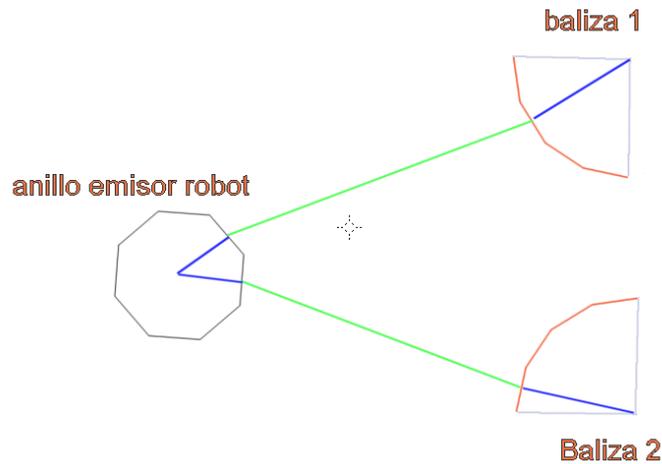
*Figura 41.Alteración de las distancias a las balizas según la disposición del robot.*

La posición de partida puede verse en la Figura 41 con la trayectoria del ultrasonido en color verde y la posición final puede verse en la Figura 42 con la trayectoria del ultrasonido en color verde y en color azul la trayectoria del sonido empleando los mismos transductores que en la posición inicial justo antes de cambiar a la trayectoria verde. En el ejemplo mostrado en la Figura 42 los ángulos de las trayectorias han sido incrementados por encima de las posibilidades físicas de los transductores para que se puedan apreciar de un modo más claro la necesidad de cambio de transductores.

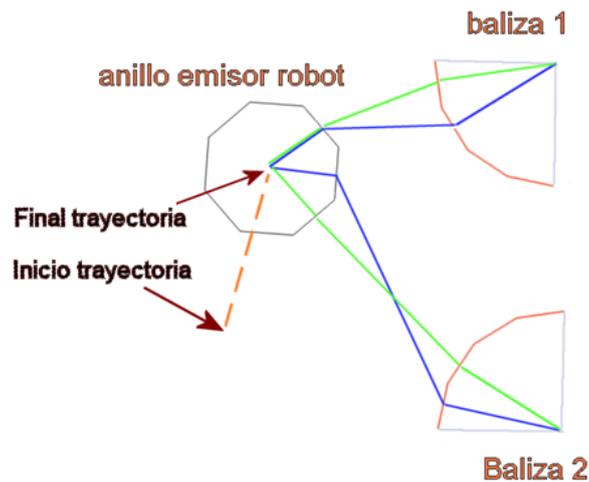
*Figura 42.Cambio de trayectorias calculadas.*

En la Figura 43 se ve la disposición del robot y la baliza con la que se han realizado las gráficas de acumulación de error por orientación. En verde se dibuja la distancia calculada por SAPPO y en azul se visualiza la distancia real. En la situación ideal de partida, el ángulo w sería de 0 grados. En este caso la distancia real es exactamente la distancia entre transductores más la apotema del anillo del robot, más la apotema del polígono de la baliza.

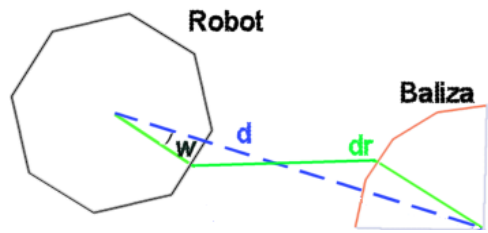

*Figura 43. Disposición de lo baliza y el anillo del robot.*

En la Figura 44 puede apreciarse la variación de la aportación del error en cm dependiendo de la orientación. Para realizar la gráfica se ha tenido en cuenta el error generado con d = 4 m constante. La gráfica muestra el incremento de distancia medida sobre la distancia real ocasionado por el incremento del ángulo de orientación entre el transductor más cercano de la baliza y del robot (nombrado con *w* en la Figura 44).

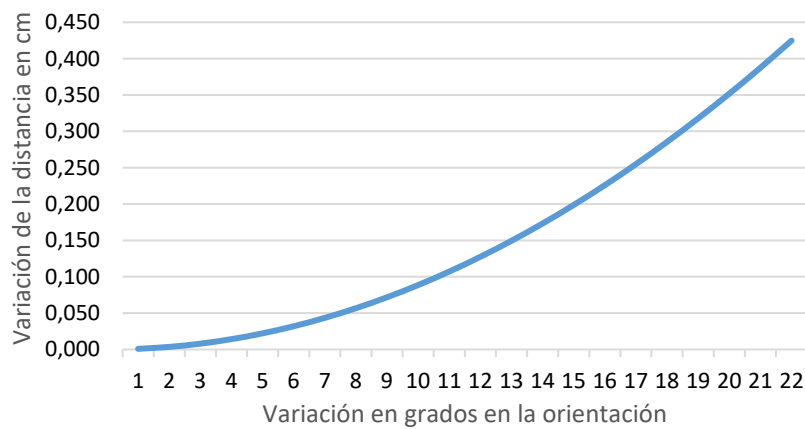

*Figura 44. Error de medida según varían los grados de desviación en la orientación.*

En el gráfico puede verse en el eje horizontal los grados de desviación entre la apotema y la línea de la trayectoria medida entre transductores. En el eje vertical se aprecia la contribución al error de la medida de distancia ocasionada por el ángulo del eje horizontal. Existen en el mercado transductores que pueden tener un ángulo de cobertura de 80 grados, pero normalmente los transductores empleados tienen un rango de cobertura de 30 grados con lo que la apotema puede tener un grado máximo de desviación que oscila entre +/- 15 grados. Empleando los transductores HC-SR04 puede apreciarse que la contribución al error con la desviación máxima de 15 grados es de 2 mm. Dado que la distancia máxima que pueden llegar a medir los transductores económicos es de 9 metros, la gráfica anterior se ha generado manteniendo una distancia media fija de 4 metros y variando la orientación del robot.

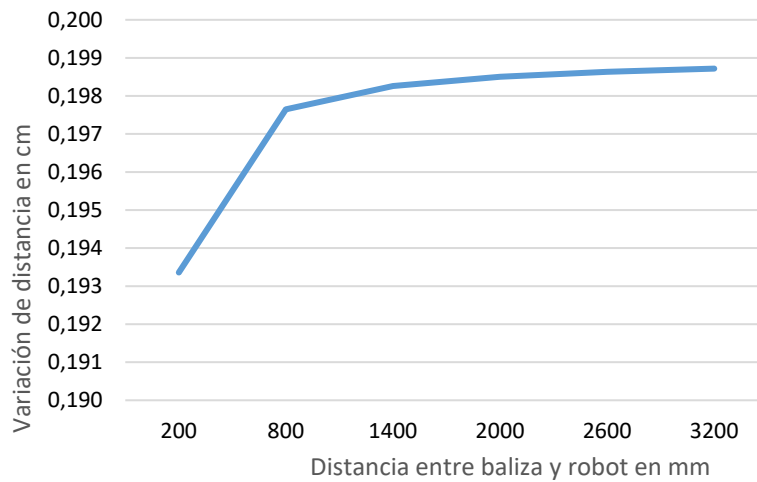
*Figura 45. Error dependiente de la distancia a 15 grados.*

Este segundo gráfico de la Figura 45 muestra cómo afecta la distancia entre los centros de los polígonos del robot y la baliza a la distribución del error de medida. El rango de contribución dependiendo de la distancia es de medio milímetro. Como puede apreciarse a medida que aumenta la distancia el aumento de la contribución es menor y a partir de 1,5 m se puede considerar nulo, lo que implica que el error de medida se mantiene consante. En esta gráfica se ha mantenido el ángulo máximo de desviación entre emisor y balizas de 15 grados y se ha ido aumentando la distancia de separación entre la baliza y el robot.

## 13. Filtrado de señales

En los mejores resultados obtenidos sin aplicar filtros se obtiene una desviación de la media de 25 µs, lo que significa que hay un error medio de 8,5 mm de media que se encuentra dentro de las especificaciones iniciales en las que se pretendía conseguir errores del rango de milímetros. Estas precisiones se consiguen casi en el 90% de las medidas. Sin embargo, el 10% restante genera medidas que alcanzan varios centímetros de imprecisión.

Si se quiere mantener una precisión milimétrica en el 100% de las medidas, es necesario filtrar la señal de salida. Debe tenerse en cuenta que la señal a filtrar se corresponde con la salida de SAPPO que es la posición en el plano que ocupa el robot y para calcular esta posición se emplean las medidas de las balizas. Estas medidas serán las señales que deben ser filtradas. Aunque es posible emplear filtros complejos como el filtro Kalman, con el que es posible realizar la operación directamente sobre la posición, empleando en su fase de estimación el modelo dinámico del robot, en este caso se empleará únicamente para el filtrado del ruido dentro en la obtención de las medidas de las balizas, dado que este modelo dinámico no forma parte del sistema de posicionamiento SAPPO, cuya función es la de obtener la posición del anillo del robot únicamente con la información de las balizas. Adicionalmente a esta aplicación de filtros iniciales a SAPPO, el propio sistema de gestión de trayectorias podría aplicar un filtro del tipo Kalman para suavizar el seguimiento de la ruta.

En los siguientes apartados se definirá la base matemática para el filtrado simple de una señal con los filtros de media móvil, el filtro EMA (*Exponential Moving Average*) y el filtro de Kalman. Otros filtros complejos muy empleados para el filtrado de señales son los filtros filtro IIR (*Infinite Impulse Response*) y el filtro FIR (*Finite Impulse Response*). Estos filtros son muy empleados para el filtrado de señales de audio, dado que pueden ser configurados para eliminar las partes de la señal que se corresponden con frecuencias que no deben formar parte de la misma. Para configurarlos se debe conocer con mucha precisión las características de la señal que debe ser filtrada. En nuestro caso la señal no sigue ningún patrón, dado que las variaciones de las medidas, e incluso la frecuencia en tiempo de estas es totalmente aleatoria y depende de condiciones impredecibles, como las posiciones relativas del robot y las balizas, las colisiones de los paquetes de comunicaciones, la respuesta de los transceptores, la generación de interrupciones tanto en el hardware del robot como de las balizas, el ruido de las señales de radio y de ultrasonidos, etc. Por lo que se considera que todos estos factores unidos formarán parte del ruido generado en las señales. Una vez estudiadas estas alteraciones se comprueba de la variabilidad de los tiempos es similar a una distribución normal de media cero, con lo que resulta válido el empleo de filtros que emplean la media de las señales como base o que intentan averiguar el valor medio de la señal minimizando el error de la desviación. Por este motivo se han seleccionado estos tres filtros para procesar la salida de SAPPO.

### 13.1. Filtro Media móvil

$$\vec{X} = \frac{\sum_{i=1}^{n} X_i}{n} \qquad (26)$$

La media móvil es uno de los filtros más sencillos de implementar, aunque tiene un consumo de memoria superior a otros, debido a que es necesario almacenar la información de los n valores de la señal de entrada anteriores. En este filtro se realiza la media continua de los últimos n valores. Tiene una respuesta bastante rápida cuando *n* es bajo, pero como contrapartida se ve muy influenciado por valores muy distantes de la media.

### 13.2. Filtro EMA

La formulación del filtro EMA se muestra en la Ecuación (27).

$$S(k) = \begin{cases} Y(0), & k = 0 \\ \alpha Y(k) + (1-\alpha)S(k-1), & k > 0 \end{cases} \qquad (27)$$

Donde $\alpha$ es el coeficiente de suavizado entre 0 y 1, *Y(k)* es el valor de la señal del sensor actual en el instante *k*, *S(k)* es el valor EMA filtrado en el instante *k* y *S(k-1)* es el valor EMA filtrado en el instante anterior *k-1*.

Este filtro tiene varias ventajas, como el emplear únicamente en memoria el valor del cálculo de la señal filtrada anterior. Mediante la asignación de distintos valores de $\alpha$ se controla la suavidad de la señal. Cuanto más bajo, mayor será la suavidad de la señal de salida y la inmunidad al ruido, pero en contraposición, la señal se verá más retrasada en el tiempo, con lo que tardará más en responder a las fluctuaciones o cambios reales de la señal de entrada.

### 13.3. Filtro de Kalman

El filtro de Kalman es un algoritmo que permite estimar variables ocultas, no observables, a partir de variables que pueden ser medidas con cierto error.

La formulación para sistemas estáticos del filtro de Kalman se muestra en las Ecuaciones (28) a (32). Inicialmente SAPPO está diseñado para ser empleado en sistemas de navegación de robots, pero dado que puede ser empleado en robots con cualquier formulación cinemática, el filtro se empleará únicamente para eliminar el ruido de las medidas de distancia de las balizas.

$$\hat{x}_{n,n} = \hat{x}_{n,n-1} + K_n(z_n - \hat{x}_{n,n-1}) \tag{28}$$

$$\hat{x}_{n+1,n} = \hat{x}_{n,n} \tag{29}$$

$$K_n = \frac{P_{n,n-1}}{P_{n,n-1} - r_n} \tag{30}$$

$$P_{n,n} = (1 - K_n)P_{n,n-1} \tag{31}$$

$$P_{n+1,n} = P_{n,n} + q_n \tag{32}$$

El filtro tiene los siguientes parámetros: en la Ecuación (28), $z_n$ es el resultado de la medición, en la Ecuación (30), $r_n$ es la varianza del error de medición y en la Ecuación (32), $q_n$ es la varianza del error del proceso, ambas varianzas deben seguir una distribución normal del media cero.

La Ecuación (28) obtiene el estado actual del sistema y la Ecuación (29) realiza la predicción obteniendo el estado estimado futuro. Dado que emplearemos el filtro para eliminar el ruido en las medidas, no aplicaremos ninguna formulación para predecir el estado futuro, por lo que el estado futuro se iguala al estado actual.

En cada iteración, el algoritmo de Kalman tiene que calcular el estado actual del proceso para lo que emplea la Ecuación (28). Esta ecuación necesita el estado anterior $\hat{x}_{n,n-1}$, junto con la medida $z_n$ y el valor de la ganancia de Kalman $K_n$. Los dos primeros valores son conocidos por lo que debe calcularse la ganancia de Kalman con la Ecuación (30) que necesita la varianza del error del proceso calculada en la iteración anterior $P_{n,n-1}$, junto con la varianza del error de estimación de la medida $r_n$, ambos conocidos. La predicción del estado siguiente se realiza mediante la Ecuación (29) que, dado que se está empleando el filtro para eliminar el ruido de las medidas, no tenemos la formulación de la estimación dinámica del sistema, por lo que la predicción del estado siguiente se iguala al estado actual. Finalmente se actualiza el valor de la covarianza del error del proceso $P_{n,n}$ con la Ecuación (31) para finalmente calcular la predicción de la covarianza del error del proceso $P_{n+1,n}$ en la Ecuación (32).

Una vez calculada la predicción, debe ajustarse la ganancia de Kalman, tarea que será realizada por la Ecuación (30). Para calcular la ganancia de Kalman se empleará la información de la covarianza en el instante de tiempo anterior $P_{n,n-1}$. Una vez calculada la ganancia $K_n$, se procede a actualizar el valor de la covarianza en la Ecuación (31) para finalmente estimar la covarianza futura en la Ecuación (32).

## 14. Referencias